%% file: main.tex
\documentclass[a4paper,fleqn]{cas-sc}

\usepackage[authoryear,longnamesfirst]{natbib}
\usepackage{graphicx}
\usepackage{booktabs}
\usepackage{multirow}
\usepackage{xcolor}
\usepackage{amsmath,amssymb}
\usepackage{url}
\usepackage{pifont}

\newcommand{\Pstar}{P\textsuperscript{*}}
\newcommand{\BGB}{\textsc{bgb}}
\newcommand{\THR}{\textsc{thr}}
\newcommand{\NC}{\textsc{nc}}

\begin{document}
\let\WriteBookmarks\relax
\def\floatpagepagefraction{1}
\def\textpagefraction{.001}

\shorttitle{Background handling in training-free OVSS}
\shortauthors{W. Yao et al.}

\title[mode = title]{What the Background Hides: Thresholds, Vocabularies and Collisions in Training-Free Open-Vocabulary Segmentation}
\tnotemark[1]
\tnotetext[1]{Code, configuration scripts, derived-dataset scripts, the collision-check tool and pre-registered contracts: \url{https://github.com/WendongYao/ovss-background-audit}; run outputs and per-image statistics: \url{https://doi.org/10.5281/zenodo.24338613}.}

\author[a,b]{Wendong Yao}
\credit{Conceptualization, Methodology, Software, Validation, Formal analysis, Investigation, Data curation, Writing -- original draft, Visualization}

\author[a,b]{Binhua Huang}
\credit{Writing -- review \& editing}

\author[a,b,c]{Soumyabrata Dev}
\cormark[1]
\ead{devs@tcd.ie}
\credit{Conceptualization, Supervision, Project administration, Funding acquisition, Writing -- review \& editing}

\affiliation[a]{organization={ADAPT Research Centre},
            addressline={O'Reilly Institute, Trinity College Dublin},
            city={Dublin 2},
            country={Ireland}}
\affiliation[b]{organization={School of Computer Science, University College Dublin},
            addressline={Belfield},
            city={Dublin 4},
            country={Ireland}}
\affiliation[c]{organization={School of Computer Science and Statistics, Trinity College Dublin},
            addressline={College Green},
            city={Dublin 2},
            country={Ireland}}
\cortext[1]{Corresponding author}

\begin{abstract}
Training-free open-vocabulary semantic segmentation (OVSS) built on CLIP has advanced rapidly, with methods typically ranked by mean intersection-over-union (mIoU) across standard benchmarks. On datasets with a background class, however, implementations rely on per-dataset thresholds and background word lists that are rarely documented or unified (11 of 13 surveyed methods configure per-dataset background decisions). We audit configurations and papers of 13 training-free OVSS methods, re-run seven from official codebases, and compare official settings against a strict unified protocol under pre-registered contracts. The protocol gap depends strongly on the benchmark---ranging from 22.8 mIoU points on PASCAL VOC (VOC21) to 0.1 points (indistinguishable from zero) on ADE20K---and alters pairwise method rankings. Despite large shifts in absolute mIoU, most of the reported improvement of later CLIP-only methods over SCLIP remains under the unified protocol (roughly 79\% of the published gain). Decomposing official background handling shows that it combines numeric settings and class-name files whose dominant component reverses across benchmarks, while no tested tuning-free rule transfers reliably across all datasets. Moreover, exact collisions between background words and target names---present in six of seven official COCO-Object configurations through ``bed''---make affected classes unpredictable under official post-processing, while the main per-class drops under language-model-generated vocabularies on VOC21 are largely attributable to a few generated terms. We offer reporting recommendations and release our evaluation code, frozen configurations, contracts, and a collision-check tool.
\end{abstract}

\begin{highlights}
\item Seven training-free OVSS methods re-run from official code under two protocols
\item The official-vs-unified gap ranges from 22.8 mIoU on VOC21 to 0.1 on ADE20K
\item Most reported progress over SCLIP survives the strict unified protocol
\item Which part of background handling matters most reverses across benchmarks
\item Exact name collisions with background words make classes unpredictable
\end{highlights}


\maketitle

\input{sections/01_introduction}

\input{sections/02_related_work}
\input{sections/03_audit_setup}

\input{sections/04_results}
\input{sections/05_discussion}
\input{sections/06_conclusion}

\section*{Data and code availability}
The evaluation code, the scripts that build every evaluated configuration (with the pinned commits of the official repositories), the derived-dataset scripts and merge lists, the collision-check tool, the frozen pre-registered contracts with their deviations, and the per-cell result tables are available at \url{https://github.com/WendongYao/ovss-background-audit}~\citep{yao2026code} (code under the MIT License; data and documents under CC~BY~4.0). The per-stage run outputs, including per-cell JSON files, run logs and the per-image statistics behind the bootstrap intervals, are archived on Zenodo at \url{https://doi.org/10.5281/zenodo.24338613}~\citep{yao2026data} (CC~BY~4.0). The datasets (PASCAL VOC 2012, COCO-Stuff, PASCAL Context and ADE20K), the model weights and the official method repositories are not redistributed; they are available from their original sources under their own terms.

\section*{Declaration of competing interest}
The authors declare that they have no known competing financial interests or personal relationships that could have appeared to influence the work reported in this paper.

\section*{Funding}
This research was conducted with the financial support of Science Foundation Ireland (now Taighde \'{E}ireann -- Research Ireland) under Grant Agreement No.\ 13/RC/2106\_P2 at the ADAPT SFI Research Centre at University College Dublin, and partly funded by the China Scholarship Council (CSC, Grant No.\ 202306540013). The funders had no role in study design, data collection and analysis, decision to publish, or preparation of the manuscript.

\printcredits

\section*{Declaration of generative AI and AI-assisted technologies in the manuscript preparation process}
During the preparation of this work the authors used Claude (Anthropic) in order to assist with writing and running the experiment code on cloud GPU instances, preparing figures and tables, and reviewing the language and structure of the manuscript. The authors designed the study, defined and froze all analysis contracts before computation, directed and checked every experiment, interpreted the results, and wrote the manuscript. After using this tool/service, the authors reviewed and edited the content as needed and take full responsibility for the content of the published article.

\bibliographystyle{cas-model2-names}
\bibliography{refs}

\appendix
\counterwithin{figure}{section}\counterwithin{table}{section}\counterwithin{equation}{section}
\input{sections/07_appendix}

\end{document}

%% file: sections/01_introduction.tex
\section{Introduction}\label{sec:intro}

Semantic segmentation assigns a semantic label to every pixel of an image and is a basic step towards dense scene understanding~\cite{long2015fcn}. Conventional models are trained under a closed-set assumption: they can only predict the categories in their training label set, and recognizing new ones requires new annotations and retraining. Open-vocabulary semantic segmentation (OVSS) removes this restriction by letting users specify the target categories with free-form text at inference time~\cite{li2022lseg,ghiasi2022openseg}. Most OVSS methods build on CLIP~\cite{radford2021clip}, which is trained contrastively on hundreds of millions of image–text pairs so that images and texts share one embedding space; a new category can be recognized simply by encoding its name. However, CLIP is trained to match whole images with their captions, so its patch-level features carry little spatial information. Used directly for pixel-level prediction, CLIP therefore produces poorly localized masks~\cite{wang2023sclip}.

Current OVSS models follow two main technical routes to address this limitation: training-based and training-free methods. Training-based methods learn dense visual-language alignment through pixel- or region-level supervision~\cite{li2022lseg,ghiasi2022openseg,cho2024catseg}, or train using only image-text pairs~\cite{xu2022groupvit,cha2023tcl}. However, these methods require additional data and training costs, and may be biased towards training categories. In contrast, training-free methods perform segmentation by adjusting the inference process of frozen pretrained models, without additional task-specific training or segmentation annotations~\cite{kombol2025survey}. This makes them suitable for annotation-scarce domains, such as remote sensing~\cite{li2025segearthov}. MaskCLIP pioneered the extraction of dense visual features for OVSS~\cite{zhou2022maskclip}. Subsequent methods focus on improving CLIP's final-layer feature aggregation: SCLIP introduces correlative self-attention, NACLIP adds a spatial neighborhood prior, and ClearCLIP removes residual connections and the feed-forward network (FFN) to reduce noise~\cite{wang2023sclip,hajimiri2025naclip,clearclip}. Beyond the final layer, ResCLIP uses intermediate-layer attention weights and scores to improve spatial correspondence, while SC-CLIP repairs anomaly tokens and fuses multi-layer features~\cite{resclip,scclip}. Other methods leverage external visual foundation models, such as DINO and SAM~\cite{caron2021dino,kirillov2023sam}: ProxyCLIP computes proxy attention from DINO or SAM features, while Trident combines local feature concatenation, SAM-guided global aggregation, and mask refinement~\cite{proxyclip,trident}. These methods are commonly evaluated using mean intersection-over-union (mIoU) on PASCAL VOC, PASCAL Context, COCO-Object, COCO-Stuff, Cityscapes, and ADE20K~\cite{everingham2010voc,mottaghi2014context,lin2014coco,caesar2018cocostuff,cordts2016cityscapes,zhou2017ade20k}.

However, even when background is treated as an evaluation category, methods may use different rules to decide which pixels belong to background, despite being evaluated under the same standard and mIoU metric. In official implementations, background may be determined through foreground confidence thresholds or predefined lists of background terms. Foreground category names may also be rephrased, and these choices are not always documented in detail. Among the 13 methods reviewed in our literature audit, 9 used a list of words rather than the single word "background" to represent background on VOC21 (Table~\ref{tab:litaudit}). Such variation makes it difficult to determine whether reported performance improvements arise from the methods themselves or from their background settings based on reported mIoU scores alone. To our knowledge, no study has quantified this contribution. Moreover, these variations can affect comparisons between methods: 6 of the 21 method pairs reversed their relative rankings on VOC21 when their official settings were replaced by a unified protocol (this result reflects the overall impact of protocol differences, rather than background processing alone). These choices also present challenges in practical applications. In new domains without annotated validation data, foreground confidence thresholds cannot be selected through validation. Background word lists may also conflict with target categories. Among the 7 reproduced official configurations, 6 included "bed" in their COCO-Object background word lists, even though "bed" is itself a target category in COCO-Object (Section~\ref{sec:decomp}). Such conflicts already occur on the original benchmark and may become more likely when users change the set of target categories. Therefore, controlling and fully disclosing background settings would help distinguish methodological contributions from background-related effects, enable comparisons under a common protocol, and provide a reproducible, tuning-free starting point for new domains lacking annotations. This motivates our research question: How much do dataset-specific background settings contribute to reported mIoU on benchmarks with a background class, and can they be replaced by rules requiring no per-dataset tuning?

To answer this question, we re-run seven representative training-free OVSS methods using their official code, covering four technical routes: last-layer attention modification (SCLIP, NACLIP), residual removal (ClearCLIP), intermediate-layer feature utilization (ResCLIP, SC-CLIP), and external vision foundation models (ProxyCLIP, Trident). First, we compare the reproduced results with the numbers published by the authors to verify the reliability of our experiments. We then compare each method's official setting (O) with a strictly unified protocol (P*) based on Text-grounded Contrastive Learning (TCL)~\cite{cha2023tcl}. Compared with TCL, P* is stricter in three aspects: it uses only the word "background" with a threshold of zero, ignores dataset-specific hyperparameter settings, and disables optional pixel-level post-processing such as PAMR, while retaining method-specific components. Based on P*, we restore the official numerical settings (THR), followed by the official class-name files (BGB), and then remove background words that collide with target categories (NC). This allows us to examine how different settings contribute to the reported benchmark results. Besides that, we construct derived benchmarks from PASCAL Context and COCO-Stuff with different proportions of background pixels to examine how protocol differences vary with background coverage. Finally, we define and freeze the research hypotheses, decision thresholds, and decision rules before each experiment, with any subsequent deviations appended to the contract.

Our evaluation reveals three main findings. First, the performance gap between official settings and the unified protocol varies considerably across benchmark datasets. Averaged over the seven methods, mIoU increases by 22.8 points on VOC21 when moving from P* to official settings, compared with only 2.0 points on Context-60. The gains from official background settings also increase with the proportion of background pixels, although background coverage alone cannot explain their magnitude. Second, different components dominate the gains from restoring official background settings across benchmarks. On VOC21, about 77\% of the gain comes from class-name files, while on COCO-Object, 77\% comes from numeric settings, such as probability thresholds, logit scales, and area thresholds. Third, on two derived benchmarks based on COCO-Stuff, removing background words that conflict with target categories improves average mIoU by 1.3 points. A smaller but consistent gain is also observed on the original COCO-Object benchmark. However, the tuning-free rules and LLM-generated vocabularies we test cannot yet consistently replace settings tuned for each dataset. Importantly, for the evaluated later CLIP-only methods, most of their reported improvements over SCLIP are retained under the unified protocol.

Our main contributions are as follows:

\begin{enumerate}

\item We audit the background configurations of 13 training-free OVSS methods and reproduce seven of them using their official code. Based on this, we compare their official settings with a strictly unified benchmark protocol (P$^*$) derived from TCL~\cite{cha2023tcl}, providing a consistent evaluation condition for fair comparison. We also release the code and configurations to support reproducibility (Section~\ref{sec:protocols}).

\item We use controlled experiments to separate the contributions of numeric background settings and class-name files to segmentation performance. We further investigate the effects of background pixel proportion, conflicts between background words and target categories, and tuning-free replacement rules to identify the sources of performance variation across benchmark settings (Section~\ref{sec:decomp}).

\item We develop conflict-checking rules and their implementation, which we will release as a standalone tool to identify overlaps between background words and target categories. We also provide a benchmark reporting checklist to help researchers disclose key background settings and report experimental results in a reproducible manner (Section~\ref{sec:checklist}).

\end{enumerate}

The remainder of this paper is organized as follows. Section~\ref{sec:related} reviews related work on OVSS, benchmark protocols, and background modeling. Section~\ref{sec:setup} introduces the scope of the audit, experimental settings, benchmark protocols, preregistered analysis, and reproduction procedures. Section~\ref{sec:results} presents the experimental results, focusing on how variations in benchmark protocols and background processing settings affect segmentation performance. Section~\ref{sec:discussion} discusses the transferability of background handling, a recommended reporting protocol, and the limitations of this study. Section~\ref{sec:conclusion} concludes the paper. The appendices provide supplementary results, details of the derived benchmarks, and the full preregistration contracts and their corresponding decisions.

%% file: sections/02_related_work.tex
\section{Related work}\label{sec:related}

\subsection{Training-free open-vocabulary segmentation}

Training-based OVSS method can utilize pixel-level or masked supervision to learn mapping relationship in dense vision-language, such as  LSeg, OpenSeg and CAT-Seg~\cite{li2022lseg, ghiasi2022openseg, cho2024catseg}; and is also able to run weakly supervised learning by only depending on image-text pair, such as GroupViT and TCL~\cite{xu2022groupvit, cha2023tcl}. In this paper, we define training-free method as method without the need of extra parameter optimization for segmentation tasks after pre-training; It is not equal when comparing training-free in this paper with not using pixel-annotation or freezing the backbone itself. For example, CLIP-DINOiser needs to Train the spatial refinement module while CLIP is frozen~\cite{wysoczanska2024clipdinoiser}; Talk2DINO also needs to training the mapping from text feature to DINO feature space~\cite{barsellotti2025talk2dino}.

For training-free OVSS, one type of method reuses the internal visual representations of frozen CLIP models to generate spatial predictions. For example, MaskCLIP utilizes value features to extract dense predictions, while CLIP Surgery improves spatial localization through a dual-path architecture and value–value attention correlations~\cite{zhou2022maskclip,li2023clipsurgery}. GEM takes a step further by extending self-self attention to query-query, key-key, and value-value interactions, and facilitates token clustering through iterative self-attention and regularization during inference~\cite{bousselham2024gem}. Other methods optimize attention or residual connections in the final layer, or utilize intermediate-layer features to recover spatial structure. Another line of work uses external vision models to provide spatial information: FreeDA builds visual prototypes offline, ProxyCLIP uses external feature correlations to construct proxy attention, Trident combines local features with cross-window aggregation, and CorrCLIP utilizes SAM-based regional constraints and DINO feature correlations to improve spatial consistency~\cite{barsellotti2024freeda,proxyclip,trident,zhang2025corrclip}. Complementary to these visual modifications, ITACLIP adjusts the input through image enhancement and category descriptions generated by language models, while FreeCP focuses on reducing redundant candidate categories and visual-language ambiguity~\cite{aydin2025itaclip,chen2025freecp}.

The mechanisms illustrated mainly make optimization on model structure, feature processing and input, while the technical configurations, compared with the benchmark settings when the methods are actually applied, are from the different phase. In addition, training-free does not mean that the methods are independent from offline resources, or there is no need for configuration on datasets during inferencing. Because of this, section~\ref{sec:protocols} audits benchmark configuration OVSS implemented; and section~\ref{sec:methods} introduces reproductions based on the 7 typical training-free methods' official codes.

\subsection{Evaluation protocols and fair comparison}\label{sec:protocols}
When making comparisons among open-vocabulary segmentation methods, TCL established a standardized benchmark protocol~\cite{cha2023tcl}. To be specific, TCL applies MMSegmentation's default category names, not allowing extensions or modifications depending on the target category, and resizes the shorter side of images while preserving the aspect ratio. TCL excludes dense CRF, which incurs high computational overhead, while maintaining PAMR, and analyzes its impact specifically. For datasets containing a background category, it also notes that a probability threshold can be considered. Because of this, mentioning following TCL does not mean that all methods follow the same rule for deciding the background category. In subsequent works, including SCLIP and ResCLIP, computationally expensive post-processing is not implemented in the main evaluations, while SCLIP still uses its own image resizing and sliding-window settings~\cite{wang2023sclip,resclip}.

Subsequent studies have controlled method comparisons in different aspects. NACLIP distinguished whether external pre-training knowledge is applied from whether post-processing is applied, and pointed out that some methods' hyperparameters would vary by dataset~\cite{hajimiri2025naclip}. ClearCLIP made a comparison on benchmarks where background existed and did not exist, and offered a SCLIP comparison using fixed, unmodified class names; SC-CLIP re-evaluated baselines under its stated protocol~\cite{clearclip,scclip}. These studies show that how we interpret comparisons between methods would be affected by additional pre-training priors, class names, and benchmark reproduction setups.

In addition to model realization and inference settings, existing studies also analyzed the impact of datasets, annotations, and vocabulary on benchmark performance. MESS assesses trained zero-shot segmentation models across various domains and points out that its PASCAL VOC figures vary from previous reports because it includes a background class. It also notes that associating extra Context labels with the background, as done in some studies, might produce better results than simply using a uniform 'background' label~\cite{blumenstiel2023mess}. RENOVATE focused on assigning more specific, fine-grained names to annotated visual regions rather than simply modifying the vocabulary of the entire dataset~\cite{huang2024renovate}. OpenBench reduced semantic overlap between training and testing categories, and discovered that there were already fine-tuned CLIP models whose advantages failed to transfer to the new setting under its investigations~\cite{liu2025openbench}. Šarić et al. distinguished the bottlenecks of mask generation, selection, and classification through oracle experiments, and pointed out that a substantial number of valid masks may be discarded because they are evaluated as no-object, a mask-level decision that differs from assigning a pixel-level background class~\cite{saric2025holdsback}. In Vocabulary-Free Semantic Segmentation, the predicted class names need to align with the reference labels, while Reichard et al. found that there is no universally optimal matching threshold across datasets~\cite{reichard2025vocabfree}. Kombol et al. pointed out that the definition of background varies across different methods, and an overlap may occur between background vocabulary and target classes, such as sky~\cite{kombol2025survey}.

Similar evaluation principles have been embraced in other areas: Musgrave et al. established standardized comparisons for metric learning; Oliver et al. managed implementations and data conditions in semi-supervised learning; Gulrajani \& Lopez-Paz focused on cross-domain evaluation and model selection; and Sincan et al. synchronized preprocessing, encoders, and training settings in sign-language translation~\cite{musgrave2020reality,oliver2018realistic,gulrajani2021domainbed,sincan2025gloss}. We draw on their common-condition approach to examine attribution in OVSS. 

Our audit covered 13 selected papers on OVSS methods and their reproducible official implementations, of which 12 were available for settings checking. The audit included CLIP-DINOiser, which contains a lightweight module for training, while all 7 reproductions in our paper are training-free from additional tasks. Based on the standards of this paper's audit, 11 of the 13 methods specified dataset-specific background decision settings (i.e., probability thresholds or saliency switches) in their official configurations, and 9 of the methods used multiple background words on VOC21. Of the 13 papers, 2 completely described background selection for auditing, 4 described it partially, 6 did not provide a description, and one was not applicable. SCLIP, NACLIP, and ResCLIP shared the same class name file and a 0.1 probability threshold on 3 benchmarks that included a background category. Based on our investigation, existing comparisons do not systematically quantify to what extent background processing settings could contribute to the differences in performance between official settings and a unified protocol in recent training-free OVSS methods. This paper utilizes the same group of official implementations, conducts evaluations under both official settings and unified protocols, and analyzes the contributions of numerical background settings and class name files in sequence. These studies examine trained models, change the benchmark or its names, or compile published numbers; none re-runs training-free methods from official code with background handling as the variable.

\input{tables/tab0_literature_audit}

\subsection{Background modeling and vocabularies}

For benchmarks containing a background category, models must decide how to handle pixels that do not belong to any target class. One approach is to classify a pixel as background when the prediction confidence is low across all foreground categories, as adopted by GEM on PASCAL VOC~\cite{bousselham2024gem}. Another approach is to leverage background-related textual concepts; for example, SC-CLIP employs stuff classes (such as sky and river) to represent the background~\cite{scclip}. As summarized in Table~\ref{tab:litaudit}, these strategies can also be combined. Consequently, merely detecting background words in text prompts is insufficient to identify a method's background handling mechanism. Furthermore, background can be inferred directly from visual responses. For example, FSeg-LLaVA activates regions based on detected foreground objects and derives the background response map from them without requiring any explicit background class names~\cite{zhang2026powerofprior}.

Predefining a background vocabulary becomes challenging when target categories are dynamically specified by users. To address this, Contrastive Concepts generates a group of competing concepts for each target query, which are sourced from contextual word co-occurrences in image captions and scene-related concepts prompted from language models~\cite{wysoczanska2025contrastive}. Consequently, the model not only predicts whether a pixel belongs to the target query, but also determines if it corresponds to visually or semantically confusable objects. When handling multiple query categories, the method aggregates competing concepts across all queries and filters out suggested terms that are overly similar to the target queries based on CLIP text embedding similarity. However, competing concepts can still introduce ambiguity in target regions due to spatial relationships such as occlusion and co-occurrence: for instance, the authors report a failure case where 'blanket' is introduced as a competing concept against 'bed', even though a blanket naturally covers and coexists with a bed. Contrastive Concepts doesn’t just overlook synonyms and parts; it also filters concepts by applying a CLIP similarity threshold of 0.9. However, it focuses specifically on a single query and earlier base methods. We examine if these collisions remain in the official class-name files of seven recent methods, as well as in the generated lists following both the similarity filter and an extra LLM conflict check (Sections~\ref{sec:decomp} and~\ref{sec:rules}).

Related work also shows that it is worth paying attention to the expression of class names and the redundant or ambiguous candidate classes: RENOVATE, ITACLIP, and FreeCP addressed regional naming, class descriptions, and ambiguous candidates separately~\cite{huang2024renovate,aydin2025itaclip,chen2025freecp}. Our analysis focuses on 2 types of vocabulary interactions that can be tested through controlled experiments, which differ from the broader class semantic processing illustrated earlier. For official class name files, we delete terms from the background vocabulary that overlap with current target names based on predefined lexical rules, and observe the segmentation results. For automatically generated background vocabulary, we compare the performance of the list filtered by CLIP similarity before and after language model conflict checking is applied.

%% file: tables/tab0_literature_audit.tex
\begin{table}[pos=!htbp]
\caption{Literature and configuration audit of training-free OVSS methods (official repositories and paper text, checked November 2026). Thr.: per-dataset background decision in the official configs (VOC21/COCO-Object/Context-60); \#bg: background words in the official class-name file (VOC21/COCO-Object); Ren.: class renaming or synonyms; Post.: pixel-level post-processing in the main table; Res.: resolution or window differs across datasets; Doc.: the paper text documents the background choices; Untuned: results without the dataset-specific choices are reported. \ding{51}~yes, $\circ$~partial, \ding{55}~no, ?~not verifiable from public material, --~not applicable.}\label{tab:litaudit}
\centering\scriptsize\setlength{\tabcolsep}{3pt}
\begin{tabular}{@{}lllcccccc@{}}
\toprule
Method & Venue & Thr. & \#bg & Ren. & Post. & Res. & Doc. & Untuned \\
\midrule
MaskCLIP & ECCV 2022 & none & -- & minor & opt. & \ding{51} & -- & \ding{51} \\
GEM & CVPR 2024 & 0.85 (VOC) & -- & ? & -- & \ding{55} & \ding{51} & \ding{55} \\
CLIP-DINOiser & ECCV 2024 & saliency (VOC/Object) & 1/1 & minor & -- & \ding{55} & \ding{51} & \ding{51} \\
SCLIP & ECCV 2024 & 0.1/0.1/0.1 & 26/26 & \ding{51} & -- & \ding{55} & \ding{55} & $\circ$ \\
ClearCLIP & ECCV 2024 & 0.5/0.4/0.15 & 1/1 & minor & -- & \ding{55} & $\circ$ & \ding{51} \\
ProxyCLIP & ECCV 2024 & 0.2/0.25/0.15 & 26/17 & minor & -- & \ding{51} & \ding{55} & \ding{55} \\
NACLIP & WACV 2025 & 0.1/0.1/0.1 & 26/26 & \ding{51} & PAMR & \ding{51} & \ding{55} & $\circ$ \\
SC-CLIP & IEEE TIP 2025 & 0.15/0.35/0.15 & 26/26 & \ding{51} & -- & \ding{55} & $\circ$ & \ding{55} \\
ResCLIP & CVPR 2025 & 0.1/0.1/0.1 & 26/26 & \ding{51} & -- & \ding{51} & \ding{55} & \ding{55} \\
ITACLIP & CVPR-W 2025 & 0.1/0.1/0.1 & 44/26 & \ding{51} & PAMR + closing & \ding{51} & $\circ$ & $\circ$ \\
Trident & ICCV 2025 & 0.2/0.25/0.15 & 26/17 & minor & SAM refine & \ding{51} & \ding{55} & $\circ$ \\
CorrCLIP & ICCV 2025 & 0.2/0.25/0.15 & 26/16 & \ding{51} & mask correction & \ding{51} & \ding{55} & $\circ$ \\
FreeCP & ICCV 2025 & purification thr & 26/26 & \ding{51} & -- & \ding{55} & $\circ$ & $\circ$ \\
\bottomrule
\end{tabular}
\end{table}

%% file: sections/03_audit_setup.tex
\section{Audit setup}\label{sec:setup}

\subsection{Methods and official configurations}\label{sec:methods}

We re-run seven representative training-free OVSS methods from their official repositories, selected for public availability of code and weights and coverage of the four branches in Section~\ref{sec:intro}: five \emph{CLIP-only} methods---SCLIP~\cite{wang2023sclip}, ClearCLIP~\cite{clearclip}, NACLIP~\cite{hajimiri2025naclip}, ResCLIP~\cite{resclip}, and SC-CLIP~\cite{scclip}---and two \emph{VFM-assisted} methods---ProxyCLIP~\cite{proxyclip} (using DINO ViT-B/8~\cite{caron2021dino}) and Trident~\cite{trident} (using DINO ViT-B/16 and SAM ViT-B~\cite{kirillov2023sam}). All seven methods use the OpenAI CLIP ViT-B/16 backbone~\cite{radford2021clip}.

All official codebases run under a unified software stack: PyTorch~2.1.2, CUDA~12.1, \texttt{mmcv}~2.1.0, \texttt{mmengine}~0.10.1, and \texttt{mmsegmentation}~1.2.2.\footnote{With repository-required dependencies \texttt{transformers}~4.37.2, \texttt{fairscale}~0.4.13, and \texttt{scikit-image}~0.22.0.} While most repositories specify \texttt{mmcv}~2.0.1 and \texttt{mmsegmentation}~1.1.1 (Trident specifies 2.1.0 and 1.2.2), any version sensitivity is verified directly by the reproduction gate in Section~\ref{sec:repro}. Experiments ran on RTX~4080~SUPER~(32\,GB) and RTX~3090~(24\,GB) GPUs (cross-GPU noise is quantified in Section~\ref{sec:repro}). External weights were downloaded from official sources and verified via SHA-256 checksums recorded in the released repository.

Table~\ref{tab:audit} summarizes official per-dataset settings touching background handling: ``\#var'' counts model keys (excluding the class-name file) that differ across official VOC21, VOC20, COCO-Object, and ADE20K configs; $\tau$ is the background probability threshold (\texttt{prob\_thd}), $s$ the logit scale (\texttt{logit\_scale}), $a$ the area threshold (\texttt{area\_thd}), and ``\#bg'' the count of background terms in the official class-name file. Every method varies 1--5 keys across datasets: e.g., ClearCLIP sets $\tau$ to $0.5$, $0.4$, and $0.15$ across VOC21, COCO-Object, and Context-60, while background vocabularies range from 1~term (ClearCLIP across all datasets, and all seven methods on Context-60) to 17 on COCO-Object (ProxyCLIP, Trident) and 26 on VOC21 (all methods except ClearCLIP) as well as on COCO-Object (SCLIP, NACLIP, ResCLIP, SC-CLIP).

Three method-intrinsic choices define the boundary between O and \Pstar{}:
(1)~ClearCLIP's repository base config specifies \texttt{model\_type=`vanilla'}, which we set to \texttt{ClearCLIP} (\texttt{ignore\_residual=True}) in both O and \Pstar{} to run the published method;
(2)~Trident enables SAM refinement (\texttt{--sam\_refine}) per its README main-table commands, kept enabled in both O and \Pstar{} as an architectural component; and
(3)~NACLIP defaults to 10-step PAMR, which we retain in O and disable in \Pstar{}.
We release the evaluation code, the scripts that build all official and derived configurations, derived-dataset scripts and merge lists, the collision-checking tool, frozen pre-registration contracts with deviations, and per-cell JSON and per-image statistics; code, configurations and contracts are at \url{https://github.com/WendongYao/ovss-background-audit}~\citep{yao2026code}, and run outputs and per-image statistics are archived at \url{https://doi.org/10.5281/zenodo.24338613}~\citep{yao2026data} (see Data and code availability).

\input{tables/tab1_config_audit}

\subsection{Datasets and background share}\label{sec:data}

We evaluate on five benchmarks:
\textbf{VOC21} (VOC~2012 val~\cite{everingham2010voc}, 1,449~images, 20 foreground classes plus background),
\textbf{VOC20} (same 1,449~images, 20 classes, no background),
\textbf{COCO-Object} (COCO \texttt{val2017}~\cite{lin2014coco}, 5,000~images, converted from COCO-Stuff~\cite{caesar2018cocostuff} with 80 thing classes and all stuff/unlabeled pixels merged into background),
\textbf{ADE20K} (val~\cite{zhou2017ade20k}, 2,000~images, 150 classes, no background), and
\textbf{Context-60} (PASCAL Context val~\cite{mottaghi2014context}, 5,104~images, 59 foreground classes plus background).
Annotation maps were verified file-by-file against our earlier stage with zero mismatches (5,000~COCO-Stuff label maps via an MD5 manifest, and 1,449~VOC2012 and 2,000~ADE20K validation masks against the earlier copies); for images, we verified the COCO \texttt{val2017.zip} archive MD5 and the ADE20K archive byte size. For Context-60, the official \texttt{.mat} archive and CodaLab JSON were inaccessible (HTTP~404 and 409), so we used PyTorch-Encoding's converted \texttt{val.pth} annotations (ETag-verified, loaded via a restricted unpickler, paired with VOC2012 images). This file holds 5,104~images---one fewer than the 5,105~count commonly cited---which cannot be supplemented without the official source. The missing image shifts mIoU by roughly $1/5105$, and all seven methods reproduce published Context-60 scores within $\pm 0.05$ under O (Table~\ref{tab:repro}), confirming the conversion.

For splits with a background class, we define the background pixel share at original annotation resolution over the full split as
\begin{equation}\label{eq:bgfrac}
\text{bg\_frac} = \frac{\sum \mathbb{I}[\text{label} = 0]}{\sum \mathbb{I}[\text{label} \neq 255]},
\end{equation}
where label~$0$ is background and label~$255$ is ignored. Original background shares are 73.3\% (VOC21), 70.5\% (COCO-Object), and 12.6\% (Context-60); VOC20 and ADE20K are assigned $\text{bg\_frac} = 0$ for reference only and excluded from correlation analyses.

Following the construction of COCO-Object from COCO-Stuff, we build two controlled derived sequences by merging selected classes into background (label~$0$) and dropping them from the query list, leaving images unchanged.
In the \textbf{Context sequence}, all 59 Context-60 foreground classes are ranked by descending validation pixel frequency (ties by class ID) and merged cumulatively until $\text{bg\_frac}$ first reaches 30\%, 50\%, and 70\%. Merging 3, 6, and 11~classes yields 57, 54, and 49~classes (including background) with actual background shares of 34.9\%, 53.0\%, and 70.7\% (Ctx~35\%, Ctx~53\%, Ctx~71\% in Table~\ref{tab:variants_full}). Because ranking is purely by frequency without separating stuff from objects, the merged classes include object classes such as \texttt{person}, \texttt{cat}, and \texttt{dog} alongside stuff classes.
In the \textbf{COCO-Stuff sequence}, starting from COCO-Stuff-171 (80 thing and 91 stuff classes, with unlabeled~$255$ mapped to background), only the 91 stuff classes are ranked by descending frequency and merged until $\text{bg\_frac}$ first reaches 30\% and $(30\% + 70.5\%)/2 \approx 50.2\%$, keeping remaining stuff and all 80 thing classes as queries. Merging 7 and 21~stuff classes yields 165 and 151~classes (including background) with actual background shares of 31.5\% and 50.6\% (COCO~31\% and COCO~51\%). Merge lists were frozen prior to GPU inference without consulting any predictions. Merging also reduces query-set size, an inherent confound noted in Section~\ref{sec:discussion}; full merge lists appear in Appendix~\ref{app:variants}.

\subsection{Evaluation settings}\label{sec:settings}

\begin{figure}[pos=tb]
  \centering
  \includegraphics[width=\linewidth]{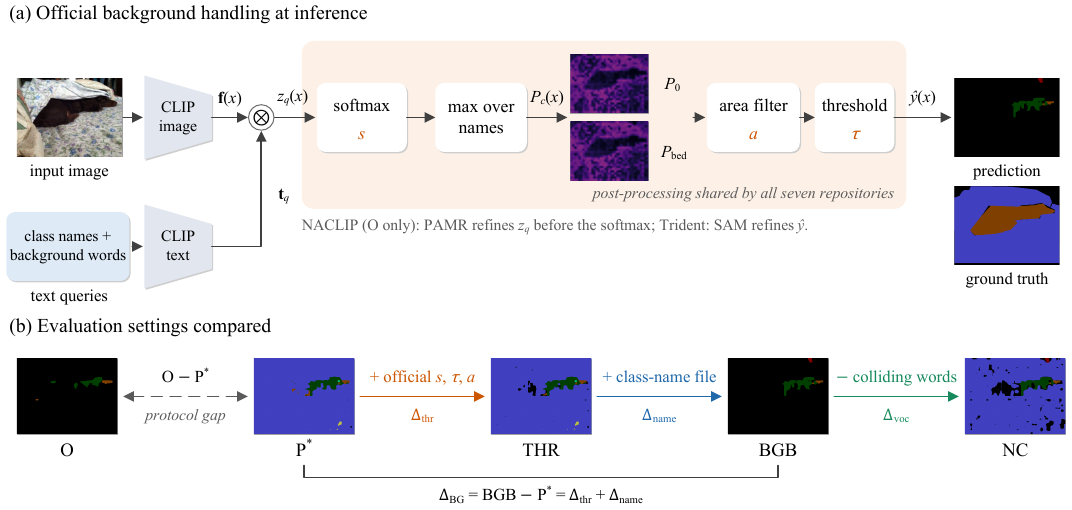}
  \caption{Overview of the evaluation settings. Thumbnails show SCLIP on one COCO-Object validation image (the 90th-percentile image of Fig.~\ref{fig:qual_bed}); colors follow the PASCAL VOC palette and heat maps show class probabilities from 0 (dark) to 1 (bright). (a)~Official background handling at inference (maps from the \BGB{} run): query similarities $z_q(x)$ are normalized by a softmax over all class and background names with logit scale $s$ (Eq.~\eqref{eq:softmax}), merged per class by a maximum (Eq.~\eqref{eq:classprob}), filtered by an area threshold $a$ (Eq.~\eqref{eq:area}) and thresholded at $\tau$ (Eq.~\eqref{eq:decision}). Because ``\texttt{bed}'' is also a background word, the background probability $P_0$ equals or exceeds the bed probability $P_{\mathrm{bed}}$ at every pixel and bed is not predicted in this image. (b)~Final predictions of the compared settings on the same image. Starting from \Pstar{}, adding the official numeric settings gives \THR{}, adding the official class-name file gives \BGB{}, and removing colliding background words gives \NC{}, so that $\Delta_{\mathrm{BG}} = \Delta_{\mathrm{thr}} + \Delta_{\mathrm{name}}$. Single-image panels are illustrative only. The diagram was drawn with AI assistance (see the declaration of generative AI use).}\label{fig:settings}
\end{figure}

Across all seven repositories, dense inference follows a shared post-processing structure that converts raw patch--text cosine similarities into pixel predictions (Fig.~\ref{fig:settings}a):
\begin{align}
p_q(x) &= \frac{\exp\bigl(s\, z_q(x)\bigr)}{\sum_{q' \in \mathcal{Q}} \exp\bigl(s\, z_{q'}(x)\bigr)}, \qquad z_q(x) = \mathbf{f}(x)^{\top}\mathbf{t}_q, \label{eq:softmax}\\
P_c(x) &= \max_{q \in \mathcal{Q}_c} p_q(x), \qquad c = 0, 1, \dots, C, \label{eq:classprob}\\
\hat{y}(x) &= \begin{cases} 0, & \max_{c} P_c(x) < \tau,\\ \arg\max_{c} P_c(x), & \text{otherwise.} \end{cases} \label{eq:decision}
\end{align}
Here, $\mathbf{f}(x)$ is the normalized visual feature at pixel~$x$ and $\mathbf{t}_q$ is the normalized text embedding of query term~$q$ (averaged across templates), so $z_q(x)$ is cosine similarity; $\mathcal{Q}_c$ collects all query terms on row~$c$ of the class-name file (with $c = 0$ denoting the background vocabulary $\mathcal{Q}_0$, and $\mathcal{Q} = \bigcup_{c=0}^{C} \mathcal{Q}_c$); $s$ is the logit scale; and $\tau$ is the background probability threshold (when $\tau = 0$, the threshold branch in Eq.~\eqref{eq:decision} never triggers because probabilities are non-negative). When an area threshold~$a$ is specified, the relative predicted area of each foreground class,
\begin{equation}
A_c = \frac{\bigl|\{x : \hat{c}(x) = c\}\bigr|}{\sum_{c'=1}^{C} \bigl|\{x : \hat{c}(x) = c'\}\bigr|}, \qquad \hat{c}(x) = \arg\max_{c} P_c(x), \label{eq:area}
\end{equation}
is computed prior to Eq.~\eqref{eq:decision}, and $P_c(x)$ is zeroed across the image for any foreground class $c \ge 1$ satisfying $A_c \le a$. Method-specific steps act outside this core: NACLIP's PAMR refines $z_q$ before Eq.~\eqref{eq:softmax}, whereas Trident's SAM refinement operates on $\hat{y}$ after Eq.~\eqref{eq:decision}. Importantly, because the softmax in Eq.~\eqref{eq:softmax} normalizes over all terms in $\mathcal{Q}$ jointly, any background term in $\mathcal{Q}_0$ competes directly with foreground queries.

We define all evaluation settings in three groups, using these short names throughout (Fig.~\ref{fig:settings}b; experimental stages are labeled V2--V5 and S1--S5 and listed in Appendix~\ref{app:contracts}).

\paragraph{Protocols.}
\begin{description}
\item[\textnormal{\textbf{O}}] (\emph{Official configuration}): each repository's per-dataset configuration used verbatim.
\item[\textnormal{\textbf{\Pstar{}}}] (\emph{Strict unified protocol, TCL-derived}): derived from TCL~\cite{cha2023tcl} to isolate background handling via seven fixed rules:
(1)~the class-name files of the ClearCLIP repository for all methods and datasets (these follow the \texttt{mmseg} names except for \texttt{ship} and \texttt{table} on VOC and joined multi-word names on ADE20K);
(2)~80 OpenAI ImageNet templates;
(3)~short side resized to 448~pixels preserving aspect ratio;
(4)~method base crop/stride, or $224/112$ when unspecified;
(5)~no pixel-level post-processing such as PAMR (architecture-intrinsic components like Trident's SAM refinement are kept);
(6)~a single ``\texttt{background}'' prompt ($\mathcal{Q}_0 = \{\text{\texttt{background}}\}$) with $\tau = 0$ and no area threshold~$a$; and
(7)~base hyperparameters only (including the base logit scale~$s$), ignoring per-dataset overrides.
Unlike TCL~\cite{cha2023tcl}, which keeps PAMR and suggests thresholding background probabilities, \Pstar{} disables both so that background handling can be cleanly isolated when comparing \BGB{}, \THR{}, and \NC{} against \Pstar{} (whereas O and \Pstar{} also differ in non-background choices such as class names, resolution, and PAMR). For ResCLIP, whose \texttt{base\_config.py} leaves required constructor parameters unset, a pre-registered fallback rule assigns keys identical across the four official configs as method-intrinsic and takes paper values (when stated) or config medians (otherwise) for varying keys.\footnote{This gives \texttt{delete\_same\_entity=True}, \texttt{attn\_rcs\_weights=[2.0, 0.5]}, \texttt{attn\_sfr\_weights=[2.1, 0.7]}, and \texttt{temp\_thd=0.18} (median of $[0.5, 0.1, 0.11, 0.25]$). No other method had missing required base-config keys.}
\end{description}

\paragraph{Decomposition of official background handling.}
\begin{description}
\item[\textnormal{\textbf{\BGB{}}}] (\emph{Transplanted official trick}): \Pstar{} with official per-dataset numeric keys ($\tau$, $s$, $a$) and official class-name files ($\mathcal{Q}_0$ and $\mathcal{Q}_c$; derived variants inherit their parent dataset's settings, appending unified COCO-Stuff names for retained stuff queries).
\item[\textnormal{\textbf{\THR{}}}] (\emph{Numeric settings only}): \Pstar{} with official numeric keys ($\tau$, $s$, $a$) only, keeping \Pstar{}'s unified class-name file ($\mathcal{Q}_0 = \{\text{\texttt{background}}\}$).
\item[\textnormal{\textbf{\NC{}}}] (\emph{Collision-free vocabulary}): \BGB{} with colliding terms deleted from the background vocabulary $\mathcal{Q}_0$ (falling back to $\{\text{\texttt{background}}\}$ if empty).
\end{description}
Averaging mIoU over the seven methods (denoted by overlines), the decomposition quantities and threshold share are defined as
\begin{equation}
\begin{aligned}
\Delta_{\mathrm{BG}} &= \overline{\BGB} - \overline{\Pstar}, &
\Delta_{\mathrm{thr}} &= \overline{\THR} - \overline{\Pstar}, &
\Delta_{\mathrm{name}} &= \overline{\BGB} - \overline{\THR}, \\
\Delta_{\mathrm{voc}} &= \overline{\NC} - \overline{\BGB}, &
\mathrm{share}_{\mathrm{thr}} &= \Delta_{\mathrm{thr}} / \Delta_{\mathrm{BG}}, &&
\end{aligned} \label{eq:decomp}
\end{equation}
where methods with zero collisions take their \BGB{} score under \NC{} (yielding $\Delta_{\mathrm{voc}} = 0$ per the S2 and S3 contracts). Thus, $\Delta_{\mathrm{BG}}$ gives the total background-handling effect, $\Delta_{\mathrm{thr}}$ isolates the numeric settings, $\Delta_{\mathrm{name}}$ isolates the official class-name file, and $\Delta_{\mathrm{voc}}$ measures the collision cost.

Because background words enter the shared softmax over $\mathcal{Q}$ in Eq.~\eqref{eq:softmax}, any background term matching a foreground query directly competes for the same pixels. Our deterministic collision rule (defining the released tool) inspects only class-name strings:
\begin{enumerate}
\item \emph{Tokenize}: split background and query lines by the method's separator (\texttt{;} for ProxyCLIP/Trident; \texttt{,} otherwise) and trim whitespace.
\item \emph{Normalize}: lowercase, convert \texttt{-} and \texttt{\_} to spaces, drop trailing ``\texttt{other}'' or ``\texttt{merged}'', and trim whitespace.
\item \emph{Match}: background word $w$ collides with query word $q$ iff $w = q$, or $w$ is a space-separated token of $q$, or $q$ is a token of $w$ (e.g., \texttt{sky} vs.\ \texttt{sky-other}; \texttt{wall} vs.\ \texttt{wall-concrete}).
\item \emph{Filter}: remove colliding words from the background line only; query lines are untouched and ``\texttt{background}'' is never removed.
\item \emph{Data independence}: execute strictly pre-inference without predictions.
\end{enumerate}

\paragraph{Tuning-free background rules.}
\begin{description}
\item[\textnormal{\textbf{\textsc{otsu}}}] (\emph{Per-image Otsu}): under \Pstar{}, computes softmax with CLIP logit scale $s = 100$ and assigns pixels whose maximum foreground probability falls below a 256-bin per-image Otsu threshold to background.\footnote{In V2-P2, \textsc{otsu} replaced \texttt{postprocess\_result}, bypassing Trident's SAM refinement; we note this wherever V2-P2 \textsc{otsu} values appear and verified in V3 via a wrapper that restoring SAM leaves conclusions unchanged.}
\item[\textnormal{\textbf{\textsc{sel26}}}] (\emph{Development-selected rule}): chosen from 120 pre-enumerated combinations (two vocabularies, $s \in \{40,\allowbreak 70,\allowbreak 100\}$, class-presence filter $r \in \{0,\allowbreak 0.2,\allowbreak 0.4,\allowbreak 0.6\}$, and five decision rules: argmax, Otsu, or fixed $t \in \{0.3,\allowbreak 0.5,\allowbreak 0.7\}$) solely on training splits (all 1,464 VOC2012 \texttt{train} images and 2,000 COCO \texttt{train2017} images, seed 20261105) across three methods (SCLIP, SC-CLIP, Trident), holding out the other four. The selected rule pairs ``\texttt{background}'' plus SCLIP's 26 VOC21 words with $s=100$, presence filter $r=0.6$ (retaining classes whose top-1\% probability mean is $\ge 0.6$ of the image maximum), and per-image Otsu. None of these components are novel; they serve strictly as an existing-heuristic baseline.
\item[\textnormal{\textbf{\textsc{cc}}}] (\emph{LLM Contrastive Concepts vocabulary}): \textsc{sel26} with the 26~words replaced by per-dataset vocabularies generated without images following Contrastive Concepts~\cite{wysoczanska2025contrastive} and filtered at CLIP cosine similarity $\beta = 0.9$. Due to 24\,GB VRAM limits, we used Qwen2.5-7B-Instruct (fp16, greedy, \texttt{max\_new\_tokens}=128) instead of Mixtral-8x7B-Instruct.
\item[\textnormal{\textbf{\textsc{cf}}}] (\emph{Conflict-filtered LLM vocabulary}): removes non-ASCII terms from \textsc{cc} and queries Qwen2.5-7B-Instruct once per word (greedy) to discard any target category, synonym, subtype, or part, without tuning on any split.
\end{description}

\subsection{Pre-registered analysis}\label{sec:prereg}

Before running the computations for each stage (V2--V5, S1--S5), we froze an analysis contract defining the questions, settings, datasets, decision metrics, numerical gates, and pre-committed conclusions (S5 being a descriptive follow-up; Appendix~\ref{app:contracts}). Contracts were never altered post hoc; hardware changes, data substitutions, and implementation discoveries were appended as dated deviations with timestamp, cause, and impact. All contracts and deviations are released and summarized in Appendix~\ref{app:contracts}.

Negative outcomes were kept and reported as findings: in V2-P2 both primary gates failed (\texttt{DISCUSS\_STOP}), showing that post-SCLIP progress mostly survives \Pstar{} and simple Otsu trails tuned tricks; in V5 the Context-60 gap missed the 5.0-mIoU gate (\texttt{STOP}), motivating our dataset-dependent framing. All transcribed tables were checked cell-by-cell against archive JSONs with zero mismatches.

To measure image-sampling uncertainty (Figs.~\ref{fig:gap} and~\ref{fig:share}), we perform image-level paired bootstrap resampling ($B = 2{,}000$, fixed per-task seeds). In each resample~$b$, mIoU is recomputed from summed intersection and union areas over non-empty-union classes as in \texttt{mmsegmentation}:
\begin{equation}
\mathrm{mIoU}^{(b)} = \frac{100}{|\mathcal{C}_b|} \sum_{c \in \mathcal{C}_b} \frac{\sum_i \omega_i^{(b)} I_{i,c}}{\sum_i \omega_i^{(b)} U_{i,c}}, \qquad \mathcal{C}_b = \Bigl\{c : \textstyle\sum_i \omega_i^{(b)} U_{i,c} > 0\Bigr\}, \label{eq:boot}
\end{equation}
where $I_{i,c}$ and $U_{i,c}$ are the intersection and union pixel counts of class~$c$ in image~$i$, and $\omega_i^{(b)}$ is the multiplicity of image~$i$ in resample~$b$. Within each dataset, all methods and settings share identical weights $\omega_i^{(b)}$, yielding 95\% percentile intervals (2.5th--97.5th percentiles) that capture image-sampling variance only, excluding GPU numerical noise.

\subsection{Reproduction}\label{sec:repro}

Before running O, we fixed reproduction targets from published main-table ViT-B/16 scores (ClearCLIP, ResCLIP, SC-CLIP, ProxyCLIP, Trident) or official READMEs (SCLIP, NACLIP), and pre-recorded all Context-60 targets. Gates required each method to stay within $\pm 1.0$~mIoU on $\ge 3/4$ benchmarks (V2-P2) and $\ge 4/7$ methods within $\pm 1.0$~mIoU on Context-60 (V5).

All seven methods pass (Table~\ref{tab:repro}): 34 of 35 cells reproduce within $\pm 0.06$~mIoU of their target. The sole exception is NACLIP on ADE20K ($18.18$ vs.\ $19.05$, $-0.87$, still within $\pm 1.0$), which we report without unverified speculation. Re-running identical settings across an RTX~4080~SUPER and RTX~3090 shifts mIoU by roughly $0.03$ ($\le 0.01$ in S1/S2 checks), whereas changing the query-word set alters \texttt{fp16} arithmetic by roughly $0.1$--$0.15$~mIoU (cosine-similarity shifts up to $3.7 \times 10^{-4}$ flipping borderline pixels and Otsu thresholds); differences below $0.15$~mIoU are therefore not interpreted. For 10~\BGB{} cells on COCO-Stuff variants in S1 exceeding 32\,GB VRAM, we used an algebraically equivalent per-class loop verified bitwise identical (\texttt{torch.equal}) up to $165 \times 200$ queries (hardware allocations and pixel-level cross-GPU checks appear in Appendix~\ref{app:repro}).

\input{tables/tab2_reproduction}

%% file: tables/tab1_config_audit.tex
\begin{table}[t]
\caption{Official per-dataset settings that touch background handling. ``\#var'' counts model keys (other than the class-name file) whose value differs across the official VOC21/VOC20/COCO-Object/ADE20K configs. $\tau$: probability threshold for background; $s$: logit scale; $a$: area threshold; ``\#bg'': number of background words in the official class-name file. ``--'': not set (class default).}\label{tab:audit}
\centering\small
\begin{tabular}{@{}lccccccccccc@{}}
\toprule
 & & \multicolumn{4}{c}{VOC21} & \multicolumn{3}{c}{COCO-Object} & \multicolumn{3}{c}{Context-60} \\
\cmidrule(lr){3-6} \cmidrule(lr){7-9} \cmidrule(lr){10-12}
Method & \#var & $\tau$ & $s$ & $a$ & \#bg & $\tau$ & $s$ & \#bg & $\tau$ & $s$ & \#bg \\
\midrule
SCLIP & 3 & 0.1 & 65 & 0.1 & 26 & 0.1 & 50 & 26 & 0.10 & 50 & 1 \\
ClearCLIP & 1 & 0.5 & -- & -- & 1 & 0.4 & -- & 1 & 0.15 & -- & 1 \\
NACLIP & 2 & 0.1 & -- & -- & 26 & 0.1 & -- & 26 & 0.10 & -- & 1 \\
ResCLIP & 5 & 0.1 & -- & -- & 26 & 0.1 & -- & 26 & 0.10 & -- & 1 \\
SC-CLIP & 3 & 0.15 & 50 & 0.1 & 26 & 0.35 & 55 & 26 & 0.15 & -- & 1 \\
ProxyCLIP & 1 & 0.2 & -- & -- & 26 & 0.25 & -- & 17 & 0.15 & -- & 1 \\
Trident & 5 & 0.2 & -- & -- & 26 & 0.25 & -- & 17 & 0.15 & -- & 1 \\
\bottomrule
\end{tabular}
\end{table}

%% file: tables/tab2_reproduction.tex
\begin{table}[t]
\caption{Official configurations (O) re-run from the official code, with the difference to the number reported by each method in parentheses. Reported targets were fixed before running.}\label{tab:repro}
\centering\small
\begin{tabular}{@{}lccccc@{}}
\toprule
Method & VOC21 & VOC20 & COCO-Object & ADE20K & Context-60 \\
\midrule
SCLIP & 59.63 ($+$0.00) & 81.53 ($-$0.01) & 33.52 ($+$0.00) & 16.45 ($+$0.00) & 31.75 ($+$0.01) \\
ClearCLIP & 51.80 ($+$0.00) & 80.93 ($+$0.03) & 33.04 ($+$0.04) & 16.65 ($-$0.05) & 32.57 ($-$0.03) \\
NACLIP & 64.06 ($-$0.01) & 83.02 ($-$0.01) & 36.20 ($+$0.01) & 18.18 ($-$0.87) & 34.99 ($+$0.00) \\
ResCLIP & 61.05 ($-$0.05) & 86.02 ($+$0.02) & 34.95 ($-$0.05) & 18.02 ($+$0.02) & 33.50 ($+$0.00) \\
SC-CLIP & 64.61 ($+$0.01) & 84.29 ($-$0.01) & 37.72 ($+$0.02) & 20.06 ($-$0.04) & 36.81 ($+$0.01) \\
ProxyCLIP & 61.24 ($-$0.06) & 80.33 ($+$0.03) & 37.54 ($+$0.04) & 20.17 ($-$0.03) & 35.30 ($+$0.00) \\
Trident & 67.08 ($-$0.02) & 84.49 ($-$0.01) & 41.07 ($-$0.03) & 21.92 ($+$0.02) & 38.65 ($+$0.05) \\
\bottomrule
\end{tabular}
\end{table}

%% file: sections/04_results.tex
\section{Results}\label{sec:results}

\subsection{Protocol gap across datasets}\label{sec:gap}

\begin{figure}[pos=tb]
  \centering
  \includegraphics[width=\linewidth]{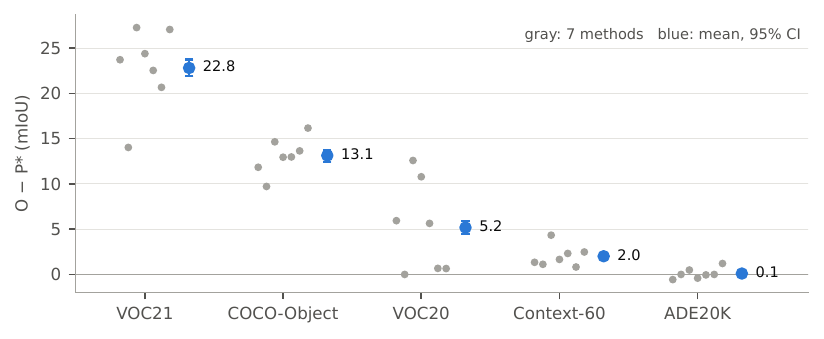}
  \caption{Protocol gap ($\text{O} - \Pstar{}$) across the five original benchmarks. Gray dots denote the seven individual methods and blue markers denote the seven-method mean with 95\% image-level paired bootstrap confidence intervals. The gap depends strongly on the benchmark, spanning from 22.8~mIoU points on VOC21 to 0.1~points (indistinguishable from zero) on ADE20K.}\label{fig:gap}
\end{figure}

The performance gap between official configurations~(O) and the strict unified protocol~(\Pstar{}) varies strongly across benchmarks, ranging from 22.8~mIoU points on VOC21 to 0.1~points on ADE20K. Averaged over the seven methods, O exceeds \Pstar{} by 22.81~points on VOC21 (95\%~CI [21.90, 23.75]; per-method range 14.03--27.27~points), whereas on ADE20K the mean gap is 0.10~points (95\%~CI [$-0.05$, 0.25]; range $-0.57$ to $+1.21$), which is not distinguishable from zero (Fig.~\ref{fig:gap} and Table~\ref{tab:main}). Across the remaining three benchmarks, O exceeds \Pstar{} by 13.14~points on COCO-Object (95\%~CI [12.47, 13.70]; range 9.72--16.17), 5.18~points on VOC20 (95\%~CI [4.49, 5.90]; range 0.00--12.59), and 2.01~points on Context-60 (95\%~CI [1.81, 2.22]; range 0.82--4.34). On Context-60, the pre-registered V5~generalization gate ($\ge 5.0$~points) was therefore not met (\texttt{STOP}), showing that large protocol gaps are not uniform across all benchmarks with a background class.

Transplanting official background handling alone (\BGB{}) recovers most of the gap on the three benchmarks that include a background class, lifting mean mIoU over \Pstar{} by 21.83~points on VOC21, 12.24~points on COCO-Object, and 1.49~points on Context-60 (Table~\ref{tab:main}). For ClearCLIP, ProxyCLIP, and Trident on VOC21, \BGB{} matches O almost exactly (51.80 vs.\ 51.80, 61.27 vs.\ 61.24, and 67.08 vs.\ 67.08), indicating that their gap arises almost exclusively from background handling. By contrast, NACLIP remains below O under \BGB{} by 7.04~points on VOC21 (57.02 vs.\ 64.06) and 3.10~points on Context-60 (31.89 vs.\ 34.99), indicating that its official configuration also relies on non-background settings (in V2 single-factor runs on VOC21, removing its default 10-step PAMR accounts for roughly 5.2~points; PAMR was not isolated separately on Context-60).

Crucially, VOC20 exhibits a mean gap of 5.18~points despite having no background class, concentrated in NACLIP (12.59~points) and ResCLIP (10.79~points). Earlier single-factor substitutions on three methods (SCLIP, ClearCLIP, NACLIP in V2) attribute roughly 3.6--3.9~points on VOC20 to class-name rewrites and 3.3~points to NACLIP's PAMR, with resolution and sliding-window changes shifting scores by at most about 2.4~points.\footnote{From V2 single-factor runs varying one setting at a time relative to each method's official package on SCLIP, ClearCLIP, and NACLIP.} Thus, O and \Pstar{} differ in non-background choices as well, which motivates isolating background handling explicitly via \BGB{}, \THR{}, and \NC{} in Section~\ref{sec:decomp}.

\begin{figure}[pos=tb]
  \centering
  \includegraphics[width=\linewidth]{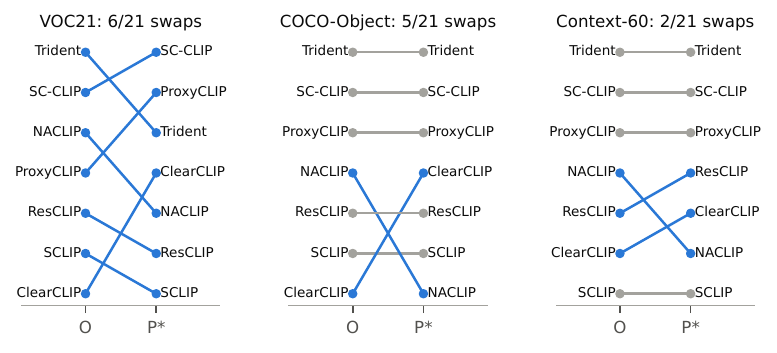}
  \caption{Method rankings under the official configuration~(O) and the strict unified protocol~(\Pstar{}) on the three benchmarks with a background class. Blue lines mark methods whose rank changes between settings. Pairwise rank swaps number 6/21 on VOC21, 5/21 on COCO-Object, and 2/21 on Context-60 (where the top three methods keep their order).}\label{fig:rank}
\end{figure}

Protocol differences also affect relative method orderings in a dataset-dependent manner. Out of 21~method pairs, pairwise rankings reverse between O and \Pstar{} for 6~pairs on VOC21, 5~on COCO-Object, and 2~on Context-60 (Fig.~\ref{fig:rank}; ADE20K likewise shows 2~swaps). On Context-60, the pre-registered V5~ranking gate ($\ge 3$ discordant pairs) was not met, and the top three methods (Trident, SC-CLIP, ProxyCLIP) retain identical orderings. Conversely, VOC20 exhibits 12~discordant pairs out of 21, underscoring that non-background configuration choices such as class-name rewrites and PAMR can reshape rankings even without a background class.

\input{tables/tab3_main_protocol}

\subsection{Relative progress under the unified protocol}\label{sec:progress}

Most of the reported improvement of later CLIP-only methods over SCLIP remains intact under the strict unified protocol~\Pstar{}. Following our pre-registered V2-P2 definition (test~T1), we measure relative progress as the mean gain over SCLIP across $\mathcal{D} = \{\text{VOC21}, \text{VOC20}, \text{COCO-Object}, \text{ADE20K}\}$:
\begin{equation}
G = \frac{1}{|\mathcal{D}|} \sum_{d \in \mathcal{D}} \Bigl( \max_{m \in \mathcal{M}'} S_{m,d} - S_{\mathrm{SCLIP},d} \Bigr), \label{eq:progress}
\end{equation}
where $\mathcal{M}' = \{\text{ClearCLIP}, \text{NACLIP}, \text{ResCLIP}, \text{SC-CLIP}\}$ and $S_{m,d}$ is the mIoU of method~$m$ on benchmark~$d$. Using published scores without heavy post-processing, this reported improvement averages $G_{\mathrm{rep}} = 5.58$~mIoU points. Under \Pstar{}, the measured improvement is $G_{\mathrm{fix}} = 4.41$~points, retaining approximately 79\% of the reported gain (with a 90th-percentile image-bootstrap bound of 4.70~points). Our pre-registered hypothesis that protocol variation accounts for at least half of the reported gain required $G_{\mathrm{fix}} \le 2.79$~points; this gate was not met (\texttt{DISCUSS\_STOP}). Thus, while absolute mIoU values shift dramatically under protocol unification, later architectural designs still outperform SCLIP under a common protocol---for example, SC-CLIP leads all CLIP-only methods on VOC21, COCO-Object, and ADE20K under \Pstar{} (Table~\ref{tab:main}). Descriptively, replacing $\mathcal{M}'$ with the two VFM-assisted methods (ProxyCLIP and Trident), which were excluded from the pre-registered CLIP-only gate by contract, yields $G_{\mathrm{rep}} = 7.13$~points versus $G_{\mathrm{fix}} = 4.96$~points.

\subsection{Effect of background share}\label{sec:share}

\begin{figure}[pos=tb]
  \centering
  \includegraphics[width=0.62\linewidth]{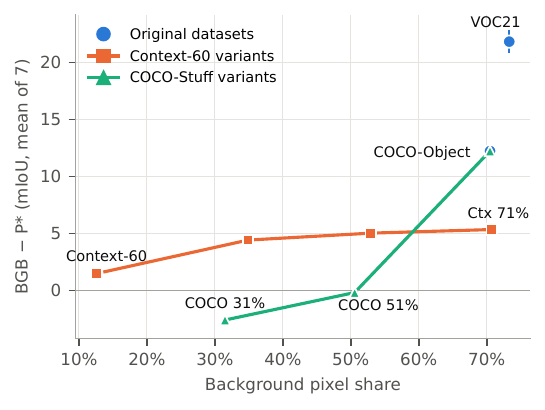}
  \caption{Background pixel share versus seven-method mean gain from official background handling (\BGB{} $-$ \Pstar{}) across original datasets and controlled derived variants; vertical bars show 95\% image-level paired bootstrap confidence intervals. Gains rise with background share overall ($\rho = 0.857$), yet diverge sharply near 71\% coverage and turn negative on COCO~31\% when stuff queries remain.}\label{fig:share}
\end{figure}

The mIoU gain from transplanted official background handling increases with background pixel share, yet background share alone does not explain the size of the gain. Across the eight evaluation splits with a background class (VOC21, COCO-Object, Context-60, and the five derived variants in Fig.~\ref{fig:share}), the pre-registered S1~test~(H1) was met (\texttt{SUPPORTED}): (a)~background share and mean $\BGB{} - \Pstar{}$ exhibit a Spearman correlation of $\rho = 0.857$ (gate $\ge 0.8$); (b)~the least-squares slope is 24.6~mIoU points per unit share with a 90\% method-level bootstrap confidence interval of [22.4, 26.9] (excluding zero; unlike the image-level 95\% intervals elsewhere, this interval resamples across the seven methods); and (c)~within the Context sequence, six of the seven methods increase monotonically from Context-60 through Ctx~35\%, Ctx~53\%, and Ctx~71\% (gate $\ge 5$), with ResCLIP as the sole exception (3.82 on Ctx~35\%, 3.74 on Ctx~53\%, and 3.44 on Ctx~71\%).

Nevertheless, matched background coverage yields widely different gains across benchmarks. Around 70--73\% background share, $\BGB{} - \Pstar{}$ averages only 5.35~points on Ctx~71\% (70.7\% background), compared with 12.24~points on COCO-Object (70.5\%) and 21.83~points on VOC21 (73.3\%). Within the Context sequence, gains rise from 1.49~points on Context-60 to 4.43~points on Ctx~35\% once high-frequency categories (\texttt{sky}, \texttt{person}, \texttt{wall}) are merged, and then level off (5.03~points on Ctx~53\%, 5.35~points on Ctx~71\%; per-method scores appear in Appendix Table~\ref{tab:variants_full}, and recall from Section~\ref{sec:data} that merging also reduces query-set size). At the per-image level within single datasets (H2, which tests $\text{O} - \Pstar{}$ rather than $\BGB{} - \Pstar{}$ across five equal-count image quintiles of background share), the pre-registered monotonicity gate ($\rho \ge 0.9$ on at least two of the three original datasets) was not met (\texttt{NOT\_SUPPORTED}): only VOC21 reached $\rho = 0.9$, while COCO-Object and Context-60 both yielded $\rho = 0.7$.

Conversely, on the COCO-Stuff variants that retain stuff classes as queries, transplanting official background handling is harmful or indistinguishable from zero. On COCO~31\%, $\BGB{} - \Pstar{}$ averages $-2.59$~points (95\%~CI [$-2.91$, $-2.27$]), with all seven methods negative; on COCO~51\%, the mean is $-0.18$~points (95\%~CI [$-0.52$, 0.16]), whose interval includes zero. A plausible explanation, not tested within S1 itself, is that official COCO-Object background lists contain terms such as \texttt{wall}, \texttt{road}, \texttt{floor}, \texttt{mountain}, \texttt{wood}, and \texttt{bed} that collide with retained query categories; we test this hypothesis directly in Section~\ref{sec:decomp}.

\subsection{Decomposition of the official background handling}\label{sec:decomp}

Official background handling combines numeric settings and class-name files whose relative contributions reverse across benchmarks, while collisions between background words and target query names impose measurable accuracy losses wherever they occur.

\begin{figure}[pos=tb]
  \centering
  \includegraphics[width=\linewidth]{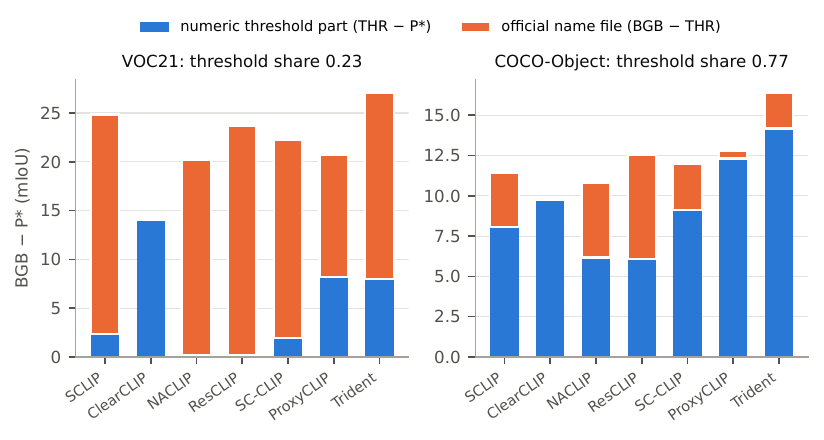}
  \caption{Decomposition of official background handling (\BGB{} $-$ \Pstar{}) into numeric settings (\THR{} $-$ \Pstar{}) and official class-name files (\BGB{} $-$ \THR{}) on VOC21 and COCO-Object. The primary driver reverses across datasets, shifting from class-name files on VOC21 (threshold share 0.23) to numeric thresholds on COCO-Object (threshold share 0.77).}\label{fig:decomp}
\end{figure}

On the original benchmarks (S3; Fig.~\ref{fig:decomp} and Appendix Table~\ref{tab:decomp}), the decomposition in Eq.~\eqref{eq:decomp} attributes the bulk of the gain to opposite components on VOC21 and COCO-Object. On VOC21, the 21.83-point gain splits into $\Delta_{\mathrm{thr}} = 4.98$~points from numeric keys ($\THR{} - \Pstar{}$) and $\Delta_{\mathrm{name}} = 16.85$~points from the official class-name file ($\BGB{} - \THR{}$), giving $\mathrm{share}_{\mathrm{thr}} = 0.23$; under our pre-registered S3 reading rule ($\le 0.33$), gains on VOC21 come mainly from the class-name file. On COCO-Object, the split reverses ($12.24 = 9.38 + 2.86$~points, $\mathrm{share}_{\mathrm{thr}} = 0.77 \ge 0.67$), so gains come mainly from numeric thresholds and scaling. Individually, numeric keys alone add almost nothing on VOC21 for several methods (NACLIP $+0.21$, ResCLIP $+0.18$). ClearCLIP provides a notable counterexample: its official file uses only ``\texttt{background}'', so its entire 14.03-point gain on VOC21 stems from $\tau = 0.5$ alone, showing that class-name dominance reflects the tested repositories' configurations rather than a universal necessity. Note that $\Delta_{\mathrm{name}}$ bundles background synonyms with foreground renamings (such as \texttt{person} and \texttt{tvmonitor} variants), which S3 does not separate further. Fig.~\ref{fig:qual} illustrates these transitions on representative validation images (further examples in Appendix~\ref{app:qual}).

Testing the collision hypothesis from Section~\ref{sec:share} on the COCO-Stuff variants (S2; Table~\ref{tab:decomp_var}), the pre-registered collision gate was met (\texttt{COLLISION\_SUPPORTED}): deleting colliding background words (\NC{} vs.\ \BGB{}) improves mIoU by $+1.30$~points averaged over all 14~method--variant pairs (ClearCLIP has zero collisions and enters the seven-method average as $0$, meeting the $\ge +1.0$ gate), and six of seven methods gain on COCO~31\% (gate $\ge 5/7$). Across the six methods with collisions (after deleting 13--15 of 26~terms in the SCLIP family and 8--9 of 17~terms in ProxyCLIP/Trident), every method improves by 1.19--1.87~points, averaging $+1.63$~points (95\%~CI [1.54, 1.71]) on COCO~31\% and $+1.39$~points (95\%~CI [1.28, 1.49]) on COCO~51\% (six-method means, whereas $+1.30$ averages all seven methods). After removing collisions, \NC{} stays within 0.3~points of \THR{} for most methods, indicating that the remaining non-colliding background words contribute little here. By contrast, the numeric component ($\THR{} - \Pstar{}$) averages $-1.15$~points on COCO~31\% (165~classes) before turning positive ($+0.94$~points) on COCO~51\% (151~classes). A plausible explanation, not tested here, is that thresholds set for 81~classes and 70.5\% background (e.g., ClearCLIP's $\tau = 0.4$ or ProxyCLIP's $0.25$) over-suppress foreground when background covers only 31.5\% of pixels. On the Ctx~71\% control, where all official files use ``\texttt{background}'' only, \THR{} and \BGB{} differed by at most 0.01~points across the three completed methods.

Finally, collisions also occur on the standard COCO-Object benchmark (S3; Appendix Table~\ref{tab:decomp}). Six of the seven official configurations include ``\texttt{bed}'' in their background list despite ``\texttt{bed}'' being one of the 80~target classes. Deleting this single colliding word improves mIoU by $+0.54$~points averaged over all seven methods (ClearCLIP scored as $0$, meeting the pre-registered S3 reading rule of $\ge +0.5$), with all six affected methods gaining $+0.57$ to $+0.73$~points.\footnote{On COCO-Object, SCLIP's collision-free \NC{} score (33.68) slightly exceeds its complete official configuration~O (33.52), though O also differs in non-background settings.} Descriptively, a follow-up mechanistic check (Appendix~\ref{app:s5}) shows that exact single-name collisions force zero predicted foreground pixels across all 92~tested cases (including \texttt{bed} on COCO-Object) and account for 70--97\% of $\Delta_{\mathrm{voc}}$ across the COCO benchmarks. Fig.~\ref{fig:qual_bed} illustrates this recovery on COCO-Object images containing \texttt{bed} (further examples in Appendix~\ref{app:qual}). Together, S2 and S3 show that while expanded background vocabularies drive most of the gain on VOC21 (where no collisions occur under our rule), colliding terms in the official COCO-Object lists degrade performance on both the COCO-Stuff variants and the original COCO-Object benchmark.

\begin{figure}[pos=tb]
  \centering
  \includegraphics[width=0.9\linewidth]{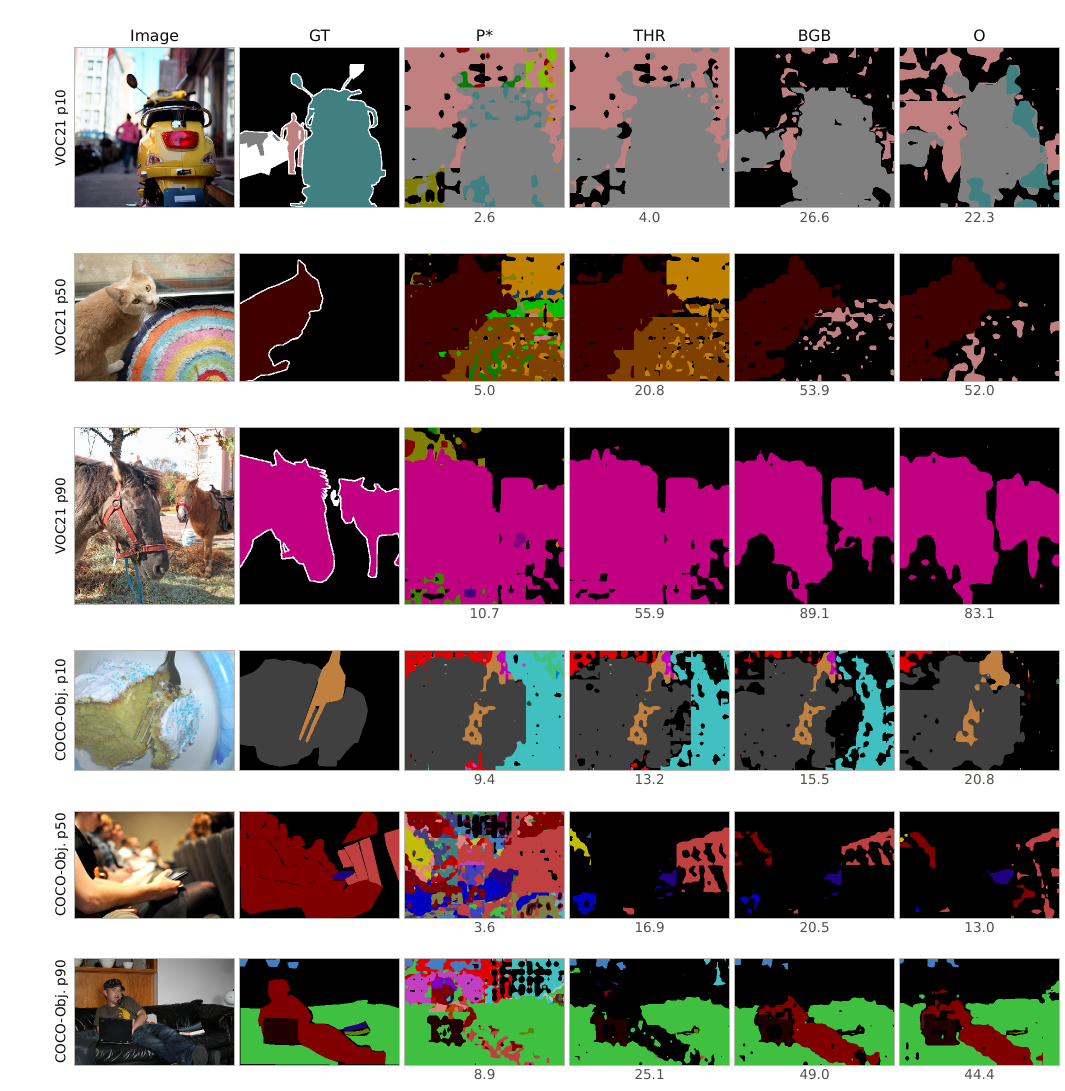}
  \caption{Qualitative SCLIP predictions under \Pstar{}, \THR{}, \BGB{} and the official configuration~(O). For each benchmark, rows are the validation images whose per-image gain $\BGB{} - \Pstar{}$ lies closest to the 10th, 50th and 90th percentiles among images with at least one foreground class and 20--80\% background pixels (VOC21: 760 of 1{,}449 images; COCO-Object: 2{,}495 of 5{,}000); selection used frozen per-image statistics only. Numbers give per-image mIoU of the original runs. Black denotes background. Images from PASCAL VOC~2012 and COCO~\texttt{val2017}.}\label{fig:qual}
\end{figure}
\begin{figure}[pos=tb]
  \centering
  \includegraphics[width=0.62\linewidth]{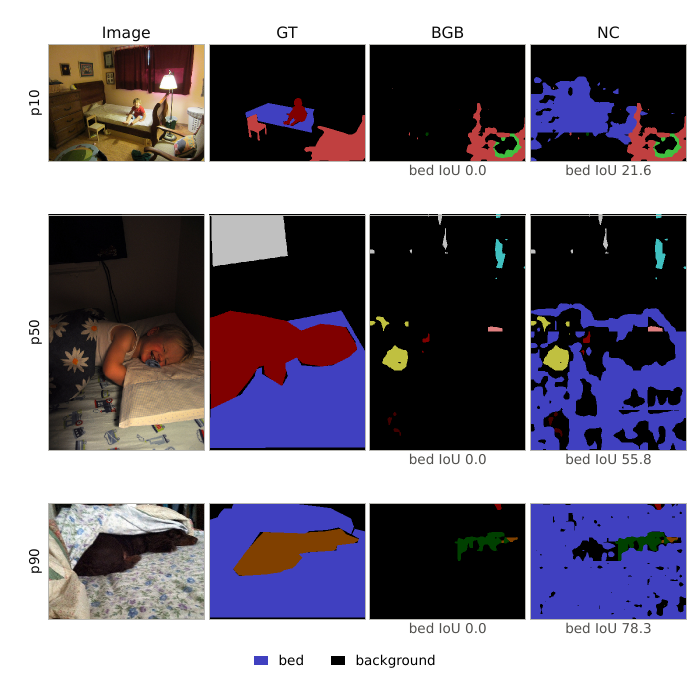}
  \caption{Effect of removing the colliding background word ``\texttt{bed}'' on COCO-Object (SCLIP). Rows are the images whose per-image bed IoU change ($\NC{} - \BGB{}$) lies closest to the 10th, 50th and 90th percentiles among the 137 validation images with at least 5\% bed pixels; selection used frozen per-image statistics only. Under \BGB{}, bed is absorbed into background (bed IoU 0 in all three images); with the colliding word removed~(\NC{}), it is recovered.}\label{fig:qual_bed}
\end{figure}
\input{tables/tab6_decomposition_variants}

\subsection{Tuning-free background rules and generated vocabularies}\label{sec:rules}

None of the tested tuning-free background rules consistently matches per-dataset official background handling across benchmarks, and LLM-generated background vocabularies suffer from collisions with foreground class names.

Simple per-image Otsu thresholding (\textsc{otsu}) improves over plain ``\texttt{background}'' (\Pstar{}) on VOC21 and COCO-Object, yet falls well behind \BGB{} by $-8.37$~points on average ($-12.15$ on VOC21 and $-4.58$ on COCO-Object; pre-registered V2-P2 gate $\ge -1.0$ not met; Table~\ref{tab:rules}). Selected solely on training splits with three methods, \textsc{sel26} matches \BGB{} closely on VOC21 ($+0.07$~points) and COCO-Object ($-0.26$~points), averaging $\Delta_{\mathrm{trick}} = -0.10$~points (meeting the V3 gate $\ge -2.0$), with the three selection methods averaging 49.91~mIoU on the training splits versus 48.87 on the validation splits and the four held-out methods averaging $+1.09$~points above \BGB{}. However, when applied unchanged to Context-60, \textsc{sel26} drops below \Pstar{} for all seven methods by $6.94$--$8.80$~points (mean $-7.90$~points below \Pstar{}, and $-9.38$ below \BGB{}). This drop is consistent with a rule selected on background-heavy benchmarks over-predicting background when background occupies only 12.6\% of pixels.

Replacing the 26 fixed words with Contrastive Concepts LLM vocabularies (\textsc{cc}, generated with Qwen2.5-7B-Instruct; Section~\ref{sec:settings}) yields $\Delta_{\mathrm{trick}}(\textsc{cc}) = -1.70$~points across VOC21 and COCO-Object. Although this passes the V4~gate ($\ge -2.0$), the two benchmarks move in opposite directions: $+0.98$~points on COCO-Object offsets $-4.37$~points on VOC21 (where VOC21 alone falls into the stop range). Adding the non-ASCII and LLM conflict filter (\textsc{cf}) across all three benchmarks gives $\Delta_{\mathrm{trick}}(\textsc{cf}) = -2.12$~points ($-3.23$ on VOC21, $+1.32$ on COCO-Object, and $-4.44$ on Context-60); the pre-registered V5~Q2 gate ($\ge -2.0$) was not met.

Per-class IoU on VOC21 (Fig.~\ref{fig:perclass}; per-method rule scores in Appendix Table~\ref{tab:rules_full}) shows that \textsc{cc} losses relative to the 26-word rule concentrate in specific object classes: \texttt{boat} ($-30.0$~points), \texttt{tvmonitor} ($-13.7$), \texttt{diningtable} ($-12.7$), \texttt{person} ($-12.2$), \texttt{bottle} ($-10.9$), \texttt{bird} ($-10.4$), and \texttt{train} ($-7.9$). On VOC21, \textsc{cf} removes only 9 of 83~terms (including \texttt{tugboat}, \texttt{feather}, and \texttt{pedestrian}), partially recovering \texttt{boat} (to $-22.1$) and \texttt{bird} (to $-2.9$), while terms such as \texttt{sailboat}, \texttt{rider}, and \texttt{remote} (associated with \texttt{boat}, \texttt{person}, and \texttt{tvmonitor}) remain unfiltered and losses on \texttt{tvmonitor}, \texttt{diningtable}, and \texttt{bottle} persist. Although individual generated terms were not ablated separately in V4/V5, this class-specific pattern is consistent with the vocabulary-collision mechanism established in Section~\ref{sec:decomp}: even after CLIP similarity filtering ($\beta = 0.9$) and an LLM conflict check, generated background lists still overlap with foreground subtypes, co-occurring objects, and parts. A subsequent leave-one-word-out ablation on the VOC21 training split (Appendix~\ref{app:s5}) isolates 10~terms whose removal on validation recovers $F = 0.70$ of the per-class drop across these seven classes, providing descriptive support for this mechanism.

\input{tables/tab4_background_rules}
\begin{figure}[pos=tb]
  \centering
  \includegraphics[width=0.62\linewidth]{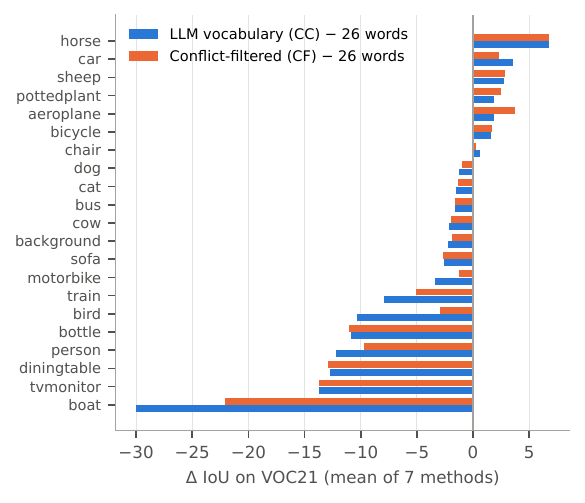}
  \caption{Per-class IoU change on VOC21 (seven-method mean) for the LLM-generated vocabulary~(CC) and conflict-filtered vocabulary~(CF) relative to the 26-word rule~(SEL26). Losses concentrate in specific classes such as \texttt{boat}, \texttt{tvmonitor}, \texttt{diningtable}, and \texttt{person}, and are only partially reduced by LLM conflict filtering.}\label{fig:perclass}
\end{figure}

\paragraph{Summary of findings.}
First, protocol unification reveals dataset-dependent gaps that alter method rankings, although most published progress of later CLIP-only methods over SCLIP survives unification. Second, official background gains scale with background coverage but are governed by different mechanisms across benchmarks---class-name files on VOC21 versus numeric thresholds on COCO-Object---and fail to transfer cleanly across datasets via tuning-free heuristics. Third, collisions between background words and target categories represent a measurable source of error in the tested official COCO configurations, and LLM-generated vocabularies exhibit class-specific drops consistent with the same mechanism. All pre-registered hypotheses, gates, and outcomes are compiled in Appendix~\ref{app:contracts}.

%% file: tables/tab3_main_protocol.tex
\begin{table*}[t]
\caption{mIoU under the official configuration (O), the strict unified protocol (P*), and P* with only the official background handling transplanted (BGB). BGB is defined only for datasets with a background class.}\label{tab:main}
\centering\scriptsize\setlength{\tabcolsep}{3pt}
\begin{tabular}{@{}lccccccccccccccc@{}}
\toprule
 & \multicolumn{3}{c}{VOC21} & \multicolumn{3}{c}{VOC20} & \multicolumn{3}{c}{COCO-Object} & \multicolumn{3}{c}{ADE20K} & \multicolumn{3}{c}{Context-60} \\
\cmidrule(lr){2-4} \cmidrule(lr){5-7} \cmidrule(lr){8-10} \cmidrule(lr){11-13} \cmidrule(lr){14-16}
Method & O & P* & BGB & O & P* & BGB & O & P* & BGB & O & P* & BGB & O & P* & BGB \\
\midrule
SCLIP & 59.63 & 35.91 & 60.76 & 81.53 & 75.59 & -- & 33.52 & 21.68 & 33.07 & 16.45 & 17.02 & -- & 31.75 & 30.41 & 31.93 \\
ClearCLIP & 51.80 & 37.77 & 51.80 & 80.93 & 80.93 & -- & 33.04 & 23.32 & 33.04 & 16.65 & 16.65 & -- & 32.57 & 31.45 & 32.57 \\
NACLIP & 64.06 & 36.79 & 57.02 & 83.02 & 70.43 & -- & 36.20 & 21.55 & 32.34 & 18.18 & 17.69 & -- & 34.99 & 30.65 & 31.89 \\
ResCLIP & 61.05 & 36.66 & 60.35 & 86.02 & 75.23 & -- & 34.95 & 22.00 & 34.56 & 18.02 & 18.43 & -- & 33.50 & 31.84 & 33.41 \\
SC-CLIP & 64.61 & 42.07 & 64.34 & 84.29 & 78.65 & -- & 37.72 & 24.74 & 36.75 & 20.06 & 20.11 & -- & 36.81 & 34.49 & 36.33 \\
ProxyCLIP & 61.24 & 40.57 & 61.27 & 80.33 & 79.66 & -- & 37.54 & 23.89 & 36.69 & 20.17 & 20.17 & -- & 35.30 & 34.48 & 35.74 \\
Trident & 67.08 & 40.02 & 67.08 & 84.49 & 83.84 & -- & 41.07 & 24.90 & 41.29 & 21.92 & 20.71 & -- & 38.65 & 36.17 & 38.03 \\
\midrule Mean & 61.35 & 38.54 & 60.37 & 82.94 & 77.76 & -- & 36.29 & 23.15 & 35.39 & 18.78 & 18.68 & -- & 34.80 & 32.78 & 34.27 \\
\bottomrule
\end{tabular}
\end{table*}

%% file: tables/tab6_decomposition_variants.tex
\begin{table*}[t]
\caption{Trick decomposition on COCO-Stuff variants that keep stuff classes as queries (background share 31\% and 51\%). NC: BGB with colliding background words removed; ``\#rm'': number of removed words; $\Delta_{\mathrm{voc}}=$NC$-$BGB. ClearCLIP's official file has no background words.}\label{tab:decomp_var}
\centering\scriptsize\setlength{\tabcolsep}{3pt}
\begin{tabular}{@{}lcccccccccccc@{}}
\toprule
 & \multicolumn{6}{c}{COCO 31\% background (165 classes)} & \multicolumn{6}{c}{COCO 51\% background (151 classes)} \\
\cmidrule(lr){2-7} \cmidrule(lr){8-13}
Method & P* & THR & BGB & NC & \#rm & $\Delta_{\mathrm{voc}}$ & P* & THR & BGB & NC & \#rm & $\Delta_{\mathrm{voc}}$ \\
\midrule
SCLIP & 21.11 & 20.99 & 19.19 & 20.73 & 15 & $+$1.54 & 20.30 & 21.75 & 20.35 & 21.71 & 13 & $+$1.36 \\
ClearCLIP & 22.64 & 18.12 & 18.12 & -- & 0 & -- & 21.73 & 19.31 & 19.31 & -- & 0 & -- \\
NACLIP & 21.54 & 21.65 & 19.91 & 21.63 & 15 & $+$1.72 & 20.35 & 22.08 & 20.81 & 22.21 & 13 & $+$1.40 \\
ResCLIP & 22.67 & 24.17 & 22.53 & 24.40 & 15 & $+$1.87 & 21.34 & 24.22 & 23.31 & 24.74 & 13 & $+$1.43 \\
SC-CLIP & 23.97 & 24.39 & 22.56 & 24.39 & 15 & $+$1.83 & 22.70 & 25.21 & 23.66 & 25.27 & 13 & $+$1.61 \\
ProxyCLIP & 24.99 & 21.12 & 19.67 & 20.97 & 9 & $+$1.30 & 23.46 & 22.50 & 21.15 & 22.34 & 8 & $+$1.19 \\
Trident & 26.06 & 24.47 & 22.89 & 24.41 & 9 & $+$1.52 & 24.44 & 25.83 & 24.48 & 25.84 & 8 & $+$1.36 \\
\bottomrule
\end{tabular}
\end{table*}

%% file: tables/tab4_background_rules.tex
\begin{table}[t]
\caption{Background handling rules, mean mIoU over the seven methods. The tuning-free rules (Otsu, SEL26, CC, CF) use no per-dataset parameters; SEL26 was selected on training splits only. Otsu for Trident lacks its SAM refinement (corrected run: conclusions unchanged; see Appendix Table~\ref{tab:rules_full}). ``--'': not run. Per-method values are in Appendix Table~\ref{tab:rules_full}.}\label{tab:rules}
\centering\small
\begin{tabular}{@{}lccc@{}}
\toprule
Rule & VOC21 & COCO-Object & Context-60 \\
\midrule
P* (plain ``background'') & 38.54 & 23.15 & 32.78 \\
Otsu rule & 48.22 & 30.81 & -- \\
SEL26 (26 words + Otsu) & 60.44 & 35.13 & 24.89 \\
LLM vocabulary (CC) & 56.00 & 36.37 & -- \\
Conflict-filtered (CF) & 57.14 & 36.71 & 29.83 \\
Official trick (BGB) & 60.37 & 35.39 & 34.27 \\
\bottomrule
\end{tabular}
\end{table}

%% file: sections/05_discussion.tex
\section{Discussion}\label{sec:discussion}

\subsection{Transferability of background handling}\label{sec:transfer}

Across OVSS benchmarks, ``background'' does not correspond to a single visual concept; rather, it is a dataset-defined residual class whose semantic scope varies with the foreground label set. On VOC21, background covers everything outside 20~object categories, including stuff regions such as sky, wall, and ground. On COCO-Object, background absorbs all 91~stuff categories and unlabeled pixels, leaving only the 80~thing categories as queries. On Context-60, by contrast, common stuff categories are foreground queries themselves, so background narrows to unlabeled and infrequent residual pixels covering just 12.6\% of the split (Section~\ref{sec:data}). A background prompt or word list set to match the residual class on one benchmark can therefore overlap directly with foreground queries on another.

A plausible hypothesis---consistent with Sections~\ref{sec:gap} and~\ref{sec:share}, though not isolated across all factors---is that thing-only benchmarks require a strong background prior, whereas stuff-rich benchmarks do not. Official background handling adds 21.8~mIoU points on VOC21 and 12.2~points on COCO-Object, compared with only 1.5~points on Context-60 (Section~\ref{sec:gap}); within the Context sequence, gains jump by roughly 3~points only once high-frequency classes such as \texttt{sky}, \texttt{person}, and \texttt{wall} are merged into background (Section~\ref{sec:share}). This hypothesis is consistent with \textsc{sel26} failing on Context-60 ($-7.9$~points below \Pstar{}; Section~\ref{sec:rules}), as would occur if a rule selected on background-heavy splits over-predicted background on a stuff-rich split. It also offers a plausible, untested explanation for why the same official numeric settings (\THR{}) hurt on COCO~31\% (165~classes) yet help on COCO~51\% (151~classes; Section~\ref{sec:decomp}). Controlled ablations that transplant Context-60 thresholds separately from background prompts, or sweep thresholds across COCO-Stuff variants, would directly test this mechanism.

These findings limit plug-and-play transfer to annotation-scarce domains~\cite{li2025segearthov}. In our literature audit, 11 of 13~methods rely on dataset-specific background rules (Table~\ref{tab:litaudit}), the primary driver reverses between VOC21 and COCO-Object (Section~\ref{sec:decomp}), and no tested tuning-free rule transfers reliably across benchmarks (Section~\ref{sec:rules}). Moreover, official repositories do not always match paper descriptions---for instance, whole-image VOC inference in NACLIP and ResCLIP, or per-dataset hyperparameters in SC-CLIP despite text stating consistent values.\footnote{Repository files and commits for each case are listed in the released audit notes.} Practitioners applying training-free OVSS to new domains therefore have no universally transferable background setting; at minimum, background vocabularies should be checked against target queries and paired with an untuned \Pstar{}-style baseline.

\subsection{Recommended reporting protocol}\label{sec:checklist}

Based on Sections~\ref{sec:gap}--\ref{sec:rules} and following reporting checklists in machine learning evaluation~\cite{dodge2019showyourwork,pineau2021reproducibility}, we recommend six practices for OVSS benchmarks:

\begin{enumerate}
\item[\textbf{R1.}] \textbf{Report official configurations~(O), a strict unified protocol~(\Pstar{}), and transplanted background handling~(\BGB{}) together.} Official-versus-unified gaps range from 0.1 to 22.8~points and reverse method rankings (Section~\ref{sec:gap}), while 11/13~audited methods specify per-dataset background decisions (Table~\ref{tab:litaudit}).
\item[\textbf{R2.}] \textbf{Report numeric background parameters ($\tau$, $s$, $a$) separately from class-name files.} Their relative contributions reverse across benchmarks---class-name files supply 77\% of the gain on VOC21 whereas numeric settings supply 77\% on COCO-Object (Section~\ref{sec:decomp})---yet only 2/13~audited papers fully document background choices (Table~\ref{tab:litaudit}).
\item[\textbf{R3.}] \textbf{Run a collision check against the target class list and disclose all removed terms.} Colliding background words appear in 6/7~reproduced official COCO-Object configs (``\texttt{bed}''), cost 1.30~points on average across COCO-Stuff variants (Section~\ref{sec:decomp}), and coincide with class-specific drops in LLM vocabularies (Section~\ref{sec:rules}); our collision tool is released for this step.
\item[\textbf{R4.}] \textbf{Report benchmarks with a background class separately from background-free benchmarks, avoiding cross-type averages.} Gaps reach 13--23~points on VOC21 and COCO-Object versus 0.1~points on ADE20K and 5.2~points on VOC20 (Section~\ref{sec:gap}), so eight-benchmark averages (e.g.,~\cite{wang2023sclip}) conflate background priors with dense alignment.\footnote{We do not suggest omitting the background class when evaluating benchmarks that define one, as we did not test that metric and~\cite{blumenstiel2023mess} note that ignoring false positives on background pixels is misleading.}
\item[\textbf{R5.}] \textbf{Select background settings on a training split and document the selection rule, rather than tuning on the evaluation split.} In Section~\ref{sec:rules}, \textsc{sel26} was chosen strictly on training splits and transferred to the validation splits without apparent overfitting, whereas 6 of 13~audited papers omit background choices altogether and 4~document them only partially (Table~\ref{tab:litaudit}).
\item[\textbf{R6.}] \textbf{Release per-dataset configurations and class-name files, and specify post-processing and inference windows.} Even without a background class, VOC20 shows a 5.2-point gap and 12/21~rank swaps; in V2 single-factor runs on three methods, class-name rewrites, PAMR, and window or resolution differences each shifted VOC20 scores (Section~\ref{sec:gap}; Table~\ref{tab:litaudit}).
\end{enumerate}

\subsection{Limitations}\label{sec:limitations}

\begin{enumerate}
\item \emph{Methods and backbone scope.} We re-ran seven training-free methods using OpenAI CLIP ViT-B/16; reserve candidate CorrCLIP (MetaCLIP backbone, pre-computed region masks) and trained OVSS models were excluded. Absolute protocol gaps may vary with backbone capacity, though dataset dependence and lexical collisions arise from label definitions and vocabularies and should remain relevant across backbones (untested).
\item \emph{Data and numerical precision.} PyTorch-Encoding's Context-60 conversion contains 5,104~images instead of 5,105 (shifting mIoU by roughly $1/5105$; confirmed by 7/7~reproductions within $\pm 0.05$ in Table~\ref{tab:repro}). Moreover, our 95\% bootstrap intervals capture image-sampling variance only, excluding cross-GPU noise ($\approx 0.03$~points) and query-set \texttt{fp16} noise ($\approx 0.1$--$0.15$~points); differences below 0.15~points are left uninterpreted (Section~\ref{sec:repro}).
\item \emph{Derived-variant confounds.} Merging classes into background simultaneously reduces query-set size, so background share and class count cannot be separated in S1 (and Context merging includes object classes such as \texttt{person}, \texttt{cat}, and \texttt{dog}; Section~\ref{sec:data}). Because the eight S1 splits share underlying images and methods, H1's correlation and slope should be read descriptively rather than as independent-sample inference.
\item \emph{Decomposition and collision rules.} Our string-matching rule ignores synonyms and part-whole relations, so measured collision losses may underestimate true lexical interference (we do not claim a formal lower bound, as removing synonyms was not tested). In addition, $\BGB{} - \THR{}$ mixes background terms with foreground renamings (Section~\ref{sec:decomp}), and the S2 Ctx~71\% control completed only 3/7~methods (reported descriptively).
\item \emph{Generated vocabularies.} Rules \textsc{cc} and \textsc{cf} used Qwen2.5-7B-Instruct with one prompt rather than Mixtral-8x7B-Instruct (Section~\ref{sec:settings}); although a subsequent training-split leave-one-word-out check (Appendix~\ref{app:s5}) attributes most of the VOC21 drop to 10~terms, absolute scores in Section~\ref{sec:rules} may vary across LLMs and prompts, and the attribution set is descriptive rather than a standalone rule.
\item \emph{Adaptive study sequence and audit snapshot.} Each experimental stage was frozen under its own pre-registered contract prior to inference, yet the choice of each subsequent stage was adapted after inspecting earlier results (such as pivoting to an audit-focused paper before V5, and reframing our claim from universal protocol gaps to dataset-dependent gaps after V5's \texttt{STOP}). Individual tests are pre-registered, whereas the overall sequence is adaptive. Finally, our literature audit reflects repositories and papers checked in November~2026.
\end{enumerate}

%% file: sections/06_conclusion.tex
\section{Conclusion}\label{sec:conclusion}

This paper audited the configurations and documentation of 13~methods from the training-free OVSS literature, re-ran seven from their official codebases, and compared their official settings against a strict unified protocol derived from TCL under pre-registered analysis contracts. We found that the protocol gap depends strongly on the benchmark---ranging from large shifts on VOC21 and COCO-Object to a modest gap on Context-60 and a near-zero difference on ADE20K---and can alter pairwise method rankings. Decomposing official background handling showed that it combines numeric settings and class-name files whose dominant component reverses across benchmarks, while none of the tested tuning-free rules transfers reliably across all datasets. We also showed that collisions between background words and target category names cause measurable accuracy losses in the tested official COCO configurations, with exact collisions making the affected class unpredictable, and that class-specific drops observed on VOC21 with LLM-generated vocabularies are largely attributable to a few generated terms that name subtypes, parts, or co-occurring objects. Still, despite substantial shifts in absolute mIoU, most of the reported improvement of later CLIP-only methods over SCLIP remains under the unified protocol. To keep methodological progress separable from background priors, we encourage the research community to adopt the six reporting recommendations~(R1--R6) in Section~\ref{sec:checklist}. Our evaluation code, official and derived configurations, collision-check tool, frozen contracts, and per-image statistics are released with this paper (see Data and code availability). Extending this audit to additional backbones and trained models while treating background handling as an explicit design choice remains worthwhile future work.

%% file: sections/07_appendix.tex
\clearpage
\section{Additional results}\label{app:results}
\suppressfloats[t]

This appendix provides the per-method breakdowns underlying Section~\ref{sec:results}: Table~\ref{tab:rules_full} reports per-method mIoU for all background-handling rules in Table~\ref{tab:rules}, Table~\ref{tab:variants_full} reports \Pstar{} and \BGB{} on the five controlled derived variants in Section~\ref{sec:share}, and Table~\ref{tab:decomp} reports the per-method decomposition of official background handling on VOC21 and COCO-Object in Section~\ref{sec:decomp}.

\input{tables/tabA1_background_rules_full}
\input{tables/tabA2_variants_full}
\input{tables/tab5_decomposition_original}

\clearpage
\section{Derived datasets}\label{app:variants}
\suppressfloats[t]

This appendix lists the classes merged into background for the controlled derived datasets used in Section~\ref{sec:data}. Following the construction rule defined there, foreground classes are ranked in descending order of validation pixel frequency (with COCO variants restricted to stuff classes) and merged cumulatively into label~$0$ until $\text{bg\_frac}$ first reaches each target. All merge lists were frozen prior to GPU inference in \texttt{tier1\_bgfrac\_frozen.json} (SHA-256 \texttt{cc5d61b6...}) without consulting any model predictions. Table~\ref{tab:merge} records the resulting background pixel shares, total class counts (including background), and merged classes. Notably, the Context sequence merges object categories (\texttt{person}, \texttt{cat}, and \texttt{dog}) alongside stuff categories, and COCO~51\% merges the stuff category \texttt{table}.

\begin{table}[pos=htb]
\centering
\caption{Classes merged into background for the derived datasets (frozen before inference). Classes are merged cumulatively in descending order of validation pixel frequency; COCO variants merge stuff classes only.}\label{tab:merge}
\small
\begin{tabular}{@{}lccp{0.52\linewidth}@{}}
\toprule
Variant & $\mathrm{bg\_frac}$ & \#classes & Merged classes (cumulative) \\
\midrule
Ctx~35\% & 0.349 & 57 & \texttt{sky}, \texttt{person}, \texttt{wall} \\
Ctx~53\% & 0.530 & 54 & $+$ \texttt{grass}, \texttt{ground}, \texttt{tree} \\
Ctx~71\% & 0.707 & 49 & $+$ \texttt{building}, \texttt{cat}, \texttt{floor}, \texttt{dog}, \texttt{water} \\
\midrule
COCO~31\% & 0.315 & 165 & \texttt{sky-other}, \texttt{tree}, \texttt{wall-concrete}, \texttt{grass}, \texttt{building-other}, \texttt{pavement}, \texttt{sea} \\
COCO~51\% & 0.506 & 151 & $+$ \texttt{road}, \texttt{clouds}, \texttt{playingfield}, \texttt{snow}, \texttt{wall-other}, \texttt{furniture-other}, \texttt{metal}, \texttt{dirt}, \texttt{table}, \texttt{sand}, \texttt{wall-tile}, \texttt{window-other}, \texttt{ceiling-other}, \texttt{fence} \\
\bottomrule
\end{tabular}
\end{table}

\clearpage
\section{Pre-registered contracts and decisions}\label{app:contracts}
\suppressfloats[t]

This appendix summarizes the analysis contracts, frozen decision rules, outcomes, and dated deviations across all experimental stages introduced in Section~\ref{sec:prereg} (Table~\ref{tab:contracts}). Each pre-registered stage froze a contract before GPU inference; results were never modified post hoc, and any hardware migrations, data-source substitutions, or implementation discoveries were appended with timestamps. Negative outcomes (\texttt{DISCUSS\_STOP} in V2-P2, Q1 \texttt{STOP} in V5, and H2 \texttt{NOT\_SUPPORTED} in S1) are retained verbatim and reported as empirical findings---in particular, V5's \texttt{STOP} outcome shifted our framing from universal protocol gaps to dataset-dependent gaps rather than being overridden. Stage~S5 was initiated descriptively after an observation in S4, with all readouts fixed prior to calculation.

\begin{table*}[pos=htb]
\centering
\caption{Pre-registered stages: question, frozen decision rule, outcome, and number of dated deviations. All contracts, deviations, and result files are released.}\label{tab:contracts}
\scriptsize\setlength{\tabcolsep}{4pt}
\begin{tabular}{@{}lp{0.22\linewidth}p{0.28\linewidth}p{0.24\linewidth}c@{}}
\toprule
Stage & Question & Frozen rule (abridged) & Outcome & Dev. \\
\midrule
V2 & Do protocol differences rival method differences? & median protocol effect $\ge$ median method gap; ranking changes in $\ge 50\%$ of 12~cells & \texttt{ADVANCE\_TO\_DESIGN} & 5 \\
V2-P2 & T1: how much reported progress over SCLIP survives \Pstar{}? T2: can a tuning-free rule match tuned tricks? & T1: $G_{\mathrm{fix}} \le 0.5\,G_{\mathrm{rep}}$ (and bootstrap p90 $\le G_{\mathrm{rep}}$); T2: $\textsc{otsu}-\BGB{} \ge -1.0$ and $\textsc{otsu}-\Pstar{} \ge +3.0$ & \texttt{DISCUSS\_STOP} (T1, T2 not met) & 9 \\
V3 & Can a rule selected on training splits approach tuned tricks? & $\Delta_{\mathrm{trick}} \ge -2.0$ and $\Delta_{\mathrm{plain}} \ge +3.0$ & \texttt{ADVANCE} & 2 \\
V4 & Does the rule survive an LLM-generated vocabulary? & $\Delta_{\mathrm{trick}}(\textsc{cc}) \ge -2.0$ (\texttt{STOP} if $< -4.0$) & \texttt{PASS} ($-1.70$; VOC21 alone $-4.38$) & 4 \\
V5 & Q1: does the protocol effect replicate on Context-60? Q2: conflict-filtered vocabulary & Q1: mean$(\text{O}-\Pstar{}) \ge 5.0$ and $\ge 3/21$ swaps; Q2: $\Delta_{\mathrm{trick}}(\textsc{cf}) \ge -2.0$ & Q1 \texttt{STOP} (2.01); Q2 vocabulary-sensitive ($-2.12$) & 5 \\
S1 & Does the gain grow with background share? & H1: $\rho \ge 0.8$, slope CI excludes 0, monotone for $\ge 5/7$ methods; H2: per-image bins $\rho \ge 0.9$ on $\ge 2/3$ datasets & H1 \texttt{SUPPORTED}; H2 \texttt{NOT\_SUPPORTED} & 8 \\
S2 & Are losses on COCO variants caused by colliding words? & mean $\Delta_{\mathrm{voc}} \ge +1.0$ and $\ge 5/7$ methods positive on COCO~31\% & \texttt{COLLISION\_SUPPORTED} & 4 \\
S3 & Thresholds vs.\ class-name files on original datasets & $\mathrm{share}_{\mathrm{thr}} \ge 0.67$ / $\le 0.33$; collision loss if mean $\Delta_{\mathrm{voc}} \ge +0.5$ and $\ge 4$ methods positive & VOC21: class-name files; COCO-Object: thresholds; collision loss measurable & 2 \\
S4 & Qualitative examples (no quantitative claim) & percentile-based image selection from frozen statistics; render gate $|\Delta\text{mIoU}| \le 0.5$ & main-figure cells pass; 2 supplementary cells failed on a different GPU; identical on original GPU type & 9 \\
S5 & Mechanism of collision losses (initiated after an observation in S4) & P1: exact sole-name collisions yield zero predictions; attribution $F \ge 0.5$ / $0.2$ & P1 holds (92/92); $F = 0.70$ & 5 \\
\bottomrule
\end{tabular}
\end{table*}

\clearpage
\section{Additional qualitative results}\label{app:qual}
\begin{figure}[pos=p]
  \centering
  \includegraphics[width=\linewidth,height=0.9\textheight,keepaspectratio]{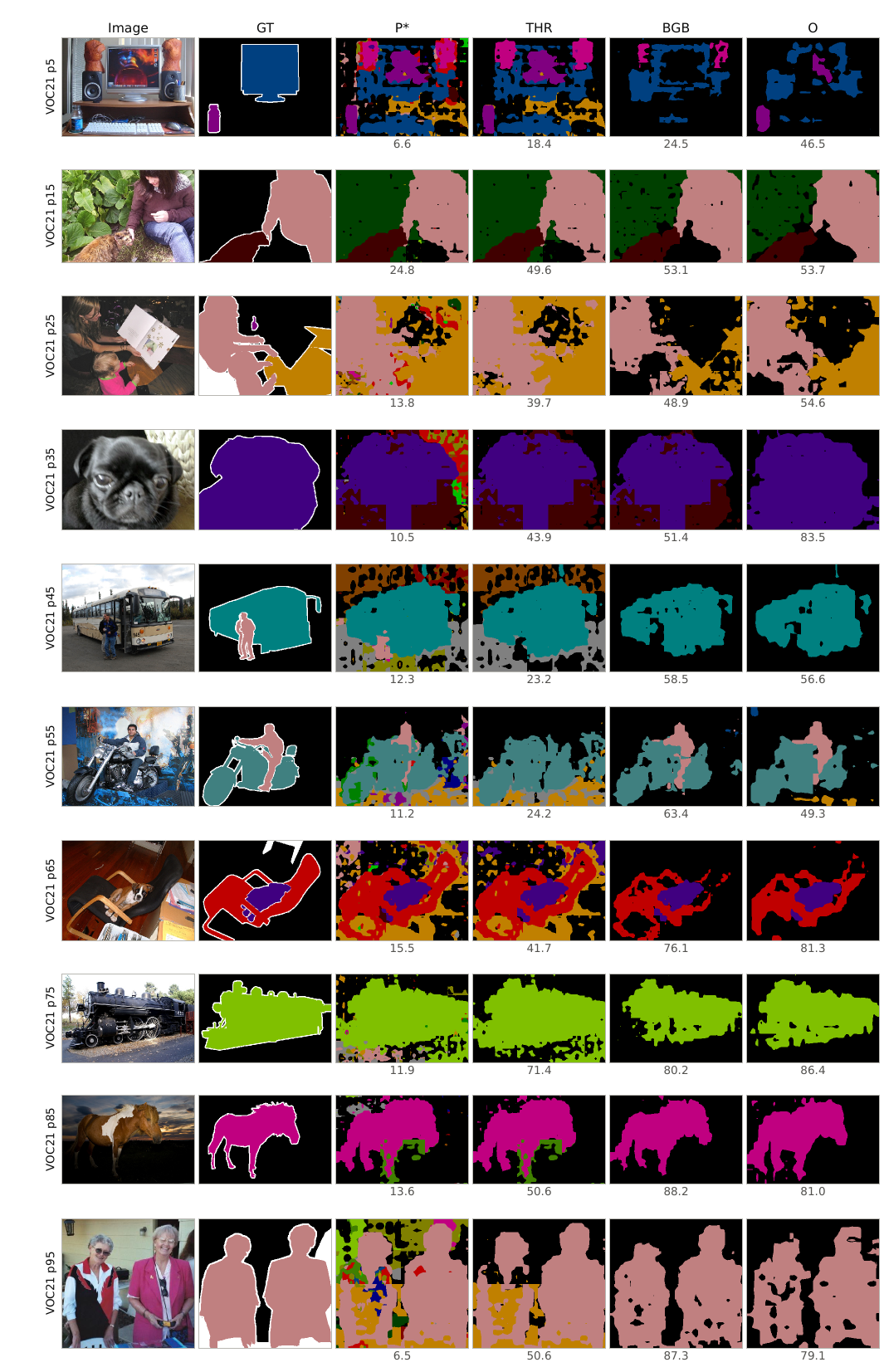}
  \caption{Additional qualitative SCLIP predictions on VOC21. Rows are the images whose per-image gain $\BGB{} - \Pstar{}$ lies closest to the 5th, 15th, \dots, 95th percentiles among eligible validation images (images shown in Fig.~\ref{fig:qual} excluded); selection used frozen per-image statistics only. Numbers give per-image mIoU of the original runs.}\label{fig:qualS_voc}
\end{figure}

\begin{figure}[pos=p]
  \centering
  \includegraphics[width=\linewidth,height=0.9\textheight,keepaspectratio]{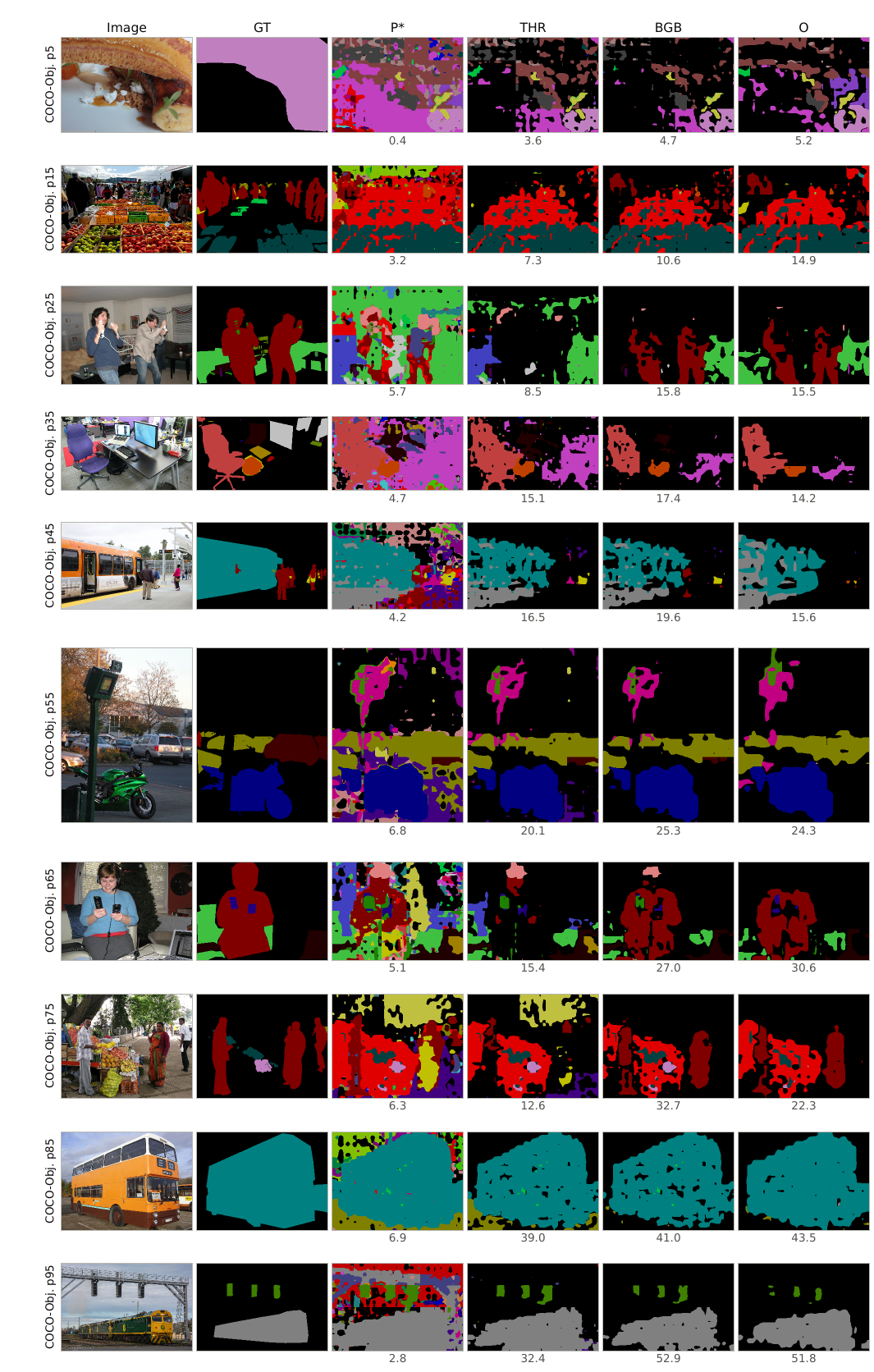}
  \caption{Additional qualitative SCLIP predictions on COCO-Object, selected as in Fig.~\ref{fig:qualS_voc}.}\label{fig:qualS_coco}
\end{figure}

\begin{figure}[pos=tb]
  \centering
  \begin{minipage}[t]{0.49\linewidth}\centering
    \includegraphics[width=\linewidth]{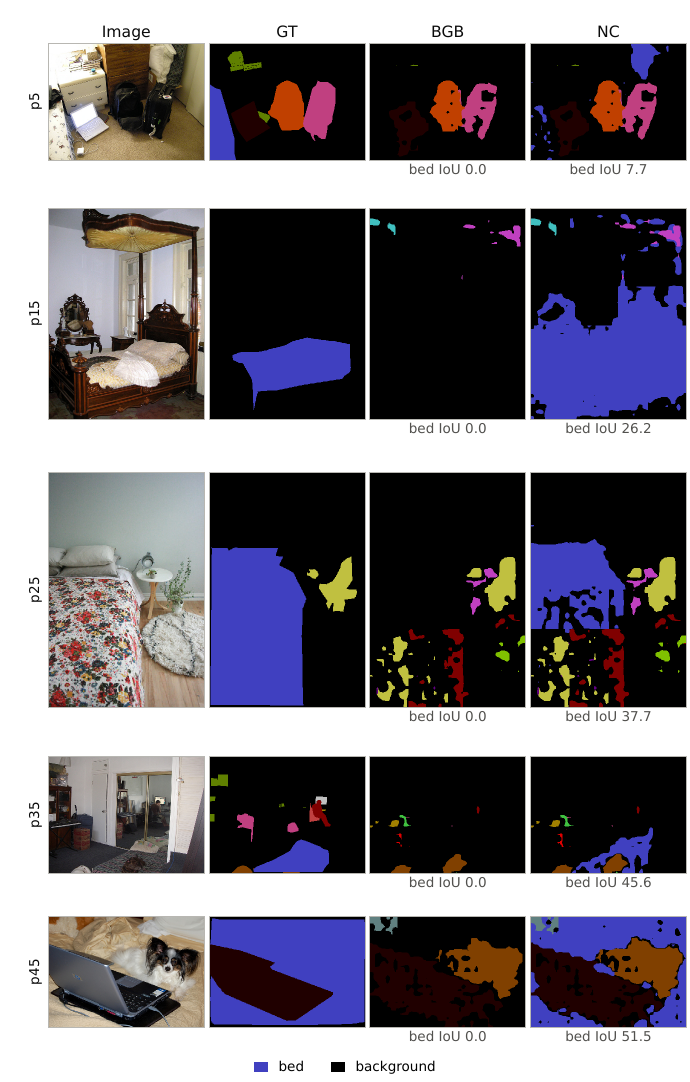}
  \end{minipage}\hfill
  \begin{minipage}[t]{0.49\linewidth}\centering
    \includegraphics[width=\linewidth]{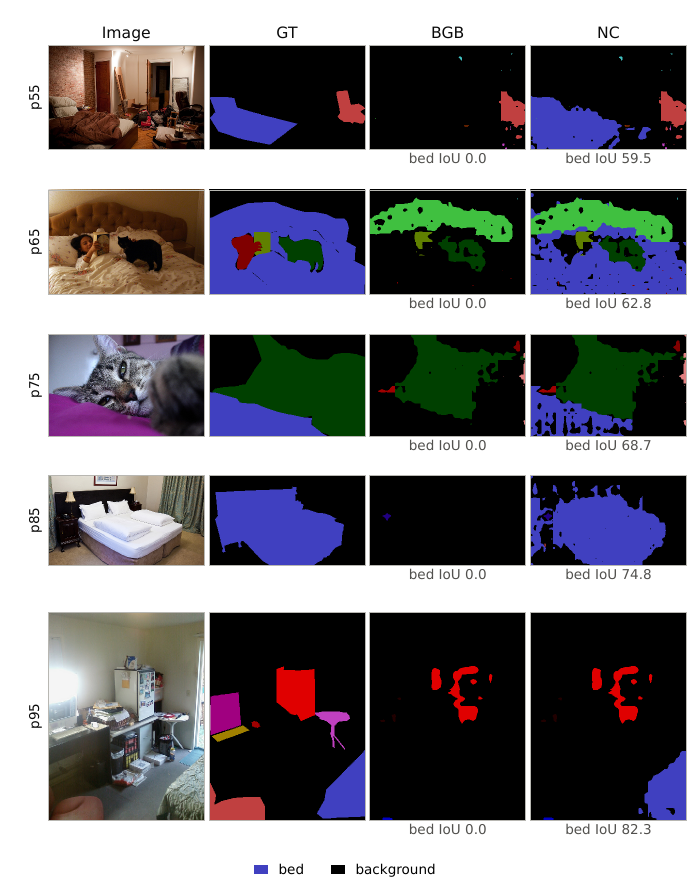}
  \end{minipage}
  \caption{Additional examples for the colliding word ``\texttt{bed}'' on COCO-Object (SCLIP): images closest to the 5th--45th (left) and 55th--95th (right) percentiles of the per-image bed IoU change $\NC{} - \BGB{}$ (images shown in Fig.~\ref{fig:qual_bed} excluded). Bed IoU is 0 under \BGB{} in all ten images.}\label{fig:qualS_bed}
\end{figure}

Figures~\ref{fig:qualS_voc}, \ref{fig:qualS_coco}, and~\ref{fig:qualS_bed} show additional qualitative SCLIP examples selected by the same deterministic percentile criteria used in Figs.~\ref{fig:qual} and~\ref{fig:qual_bed}.

\clearpage
\section{Collision mechanism}\label{app:s5}
\suppressfloats[t]

This appendix details the descriptive mechanistic follow-up~(S5) referenced in Sections~\ref{sec:decomp} and~\ref{sec:rules}; it was initiated after an observation in S4, with all readouts fixed prior to computation, and is reported descriptively without altering any pre-registered gate from V2--S4.

\paragraph{Proposition (structural zeroing under exact sole-name collisions).}
Suppose foreground class~$c \ge 1$ has a single query string $q$ ($\mathcal{Q}_c = \{q\}$) that also appears verbatim in the background vocabulary ($q \in \mathcal{Q}_0$). By Eq.~\eqref{eq:classprob},
\begin{equation}\label{eq:zeroing}
P_0(x) = \max_{q' \in \mathcal{Q}_0} p_{q'}(x) \ge p_q(x) = P_c(x)
\end{equation}
holds at every pixel~$x$. Because PyTorch's $\arg\max$ breaks ties in favor of the smallest class index ($0$ for background), class~$c$ is never selected prior to thresholding; applying $\tau$ or $a$ can only reassign pixels to class~$0$ and preserves zero predictions. For methods with post-$\arg\max$ refinement (such as Trident's SAM refinement), Eq.~\eqref{eq:zeroing} serves as an empirical expectation tested below.

\paragraph{Layer~1: verification on frozen outputs.}
Across the frozen S2 and S3 collision manifests, we classify each (method, dataset, background word, query) collision as \emph{exact sole-name} (\textbf{E-type}: verbatim string equality with $|\mathcal{Q}_c| = 1$; 92~instances across 17~dataset--class pairs and 18~cells, involving \texttt{bed}, \texttt{curtain}, \texttt{fence}, \texttt{hill}, \texttt{mountain}, \texttt{river}, \texttt{road}, \texttt{shelf}, and \texttt{wood}) or \emph{other} (\textbf{T-type}: normalized/token-level match or multi-synonym $\mathcal{Q}_c$; 188~instances). Exact collisions make a class unpredictable under the official post-processing: all 92/92 E-type instances receive 0~predicted pixels under \BGB{} (and 6/6 receive 0 under O on original COCO-Object), including Trident with SAM refinement, whereas 0/92 remain zero under \NC{}. By contrast, T-type collisions are not structurally zeroed, retaining a median predicted pixel count of approximately $0.75$ of \NC{}.

\paragraph{Layer~2: per-class decomposition of $\Delta_{\mathrm{voc}}$.}
Table~\ref{tab:s5_l2} decomposes $\Delta_{\mathrm{voc}} = \mathrm{mIoU}(\NC{}) - \mathrm{mIoU}(\BGB{})$ into contributions from E-type classes, T-type classes, and all remaining classes averaged over the six methods with collisions. E-type classes (which have IoU~0 under \BGB{} and recover to 0.2--56.0 under \NC{}, e.g., 45.0--56.0 for \texttt{bed} on COCO-Object) account for 97\% of $\Delta_{\mathrm{voc}}$ on COCO-Object and 70--71\% on the COCO-Stuff variants.

\begin{table}[pos=htb]
\centering
\caption{Decomposition of $\Delta_{\mathrm{voc}}$ (six-method mean over methods with collisions) into contributions from structurally zeroed exact sole-name collisions (E-type), other colliding classes (T-type), and remaining classes.}\label{tab:s5_l2}
\small
\begin{tabular}{@{}lccccc@{}}
\toprule
Dataset & $\Delta_{\mathrm{voc}}$ & E-type contrib. & T-type contrib. & Rest contrib. & E-type share \\
\midrule
COCO-Object & $+$0.635 & $+$0.614 & $+$0.000 & $+$0.021 & 0.97 \\
COCO~31\% & $+$1.631 & $+$1.153 & $+$0.186 & $+$0.292 & 0.71 \\
COCO~51\% & $+$1.393 & $+$0.971 & $+$0.163 & $+$0.259 & 0.70 \\
\bottomrule
\end{tabular}
\end{table}

\paragraph{Layer~3: leave-one-word-out ablation of the LLM vocabulary.}
To test whether the VOC21 per-class losses of \textsc{cc} in Section~\ref{sec:rules} can be attributed to specific generated words, we evaluated all 83~single-word removals from \textsc{cc} on the VOC21 training split (1,464~images, six methods excluding Trident due to SAM cost). Selecting words $w$ that raise any target class $c \in K = \{\text{ship}, \text{tvmonitor}, \text{table}, \text{person}, \text{bottle}, \text{bird}, \text{train}\}$ by $\ge 1.0$~IoU points on average yields a 10-word attribution set~$A$ (Table~\ref{tab:s5_l3}): subtypes (\texttt{sailboat}, \texttt{tugboat} $\to$ \texttt{ship}), parts (\texttt{feather} $\to$ \texttt{bird}, \texttt{engine} $\to$ \texttt{train}), contents or co-occurring items (\texttt{wine} $\to$ \texttt{bottle}, \texttt{napkin} $\to$ \texttt{table}, \texttt{collar} $\to$ \texttt{person}), and co-occurring scene objects (\texttt{laptop}, \texttt{tv\_stand}, \texttt{stereo} $\to$ \texttt{tvmonitor}). Removing all 10~words once on the VOC21 validation split across all seven methods recovers $F = \sum_{c \in K} R_c / \sum_{c \in K} D_c = 68.23 / 97.70 = 0.70$ of the per-class drop across $K$ (with non-$K$ classes changing by $+0.38$ on average, range $-0.07$ to $+1.22$). Because $A$ was selected using ground-truth class labels, $\textsc{cc} - A$ is reported solely as descriptive attribution and must not be interpreted as a new tuning-free rule or performance claim. Notably, none of these 10~terms match the string-collision rule in Section~\ref{sec:settings}, explaining why our lexical check does not detect subtypes, parts, or co-occurring scene objects.

\begin{table}[pos=htb]
\centering
\caption{Layer~3 leave-one-word-out attribution on VOC21 (seven-method mean on validation). $D_c = \mathrm{IoU}_c(\textsc{sel26}) - \mathrm{IoU}_c(\textsc{cc})$; $R_c = \mathrm{IoU}_c(\textsc{cc}-A) - \mathrm{IoU}_c(\textsc{cc})$. Overall recovery $F = \sum_{c \in K} R_c / \sum_{c \in K} D_c = 0.70$. Class names follow \Pstar{} (where \texttt{boat} and \texttt{diningtable} are written \texttt{ship} and \texttt{table}).}\label{tab:s5_l3}
\small
\begin{tabular}{@{}llccl@{}}
\toprule
Class $c$ & Attributed words in $A$ & $D_c$ & $R_c$ & Relation \\
\midrule
\texttt{ship} & \texttt{sailboat}, \texttt{tugboat} & 29.97 & 23.48 & subtype \\
\texttt{tvmonitor} & \texttt{laptop}, \texttt{tv\_stand}, \texttt{stereo} & 13.70 & 14.83 & scene object \\
\texttt{table} & \texttt{napkin} & 12.69 & 4.55 & co-occurring item \\
\texttt{person} & \texttt{collar} & 12.21 & 2.77 & co-occurring item \\
\texttt{bottle} & \texttt{wine} & 10.89 & 12.57 & content \\
\texttt{bird} & \texttt{feather} & 10.36 & 7.75 & part \\
\texttt{train} & \texttt{engine} & 7.89 & 2.29 & part \\
\bottomrule
\end{tabular}
\end{table}

\section{Implementation and numerical reproducibility}\label{app:repro}

This appendix documents hardware allocation and pixel-level cross-GPU reproducibility supporting Section~\ref{sec:repro}. All experiments from V2-P2 through S5 ran under PyTorch~2.1.2, CUDA~12.1, \texttt{mmcv}~2.1.0, \texttt{mmengine}~0.10.1, and \texttt{mmsegmentation}~1.2.2 (V2 ran on an RTX~3090~Ti with PyTorch~2.6.0+cu124); V2-P2, V4, and S1 used RTX~4080~SUPER~(32\,GB) GPUs, whereas V3, V5, S2, S3, and S5 ran on RTX~3090~(24\,GB) GPUs, and S4 rendered on both (RTX~4080~SUPER, then RTX~3090). Repository commits and SHA-256 checksums for weights and frozen files are recorded in the released archive.

In S4, re-rendering 39~SCLIP \THR{} and \NC{} image--setting pairs on the original RTX~3090 reproduced all 39~per-image class intersection and union counts bit-for-bit (0.000 difference). Rendering the same pairs on an RTX~4080~SUPER produced small numerical pixel flips---specifically 349 of 230,400~pixels and 204 of 307,200~pixels on two COCO-Object \THR{} images---of which 1 and 8~pixels fell on classes absent from ground truth, shifting single-image mIoU by $-0.96$ and $-1.83$~points. Such flips affect only borderline pixels; consistent with Section~\ref{sec:repro}, we do not interpret dataset-level differences below 0.15~mIoU.

%% file: tables/tabA1_background_rules_full.tex
\begin{table*}[pos=htb]
\caption{Per-method mIoU for the background rules of Table~\ref{tab:rules}. Otsu for Trident is the V2-P2 run without SAM refinement (corrected values 50.99/33.48 in V3; conclusions unchanged).}\label{tab:rules_full}
\centering\scriptsize\setlength{\tabcolsep}{2pt}
\begin{tabular}{@{}lcccccccccccccccccc@{}}
\toprule
 & \multicolumn{6}{c}{VOC21} & \multicolumn{6}{c}{COCO-Object} & \multicolumn{6}{c}{Context-60} \\
\cmidrule(lr){2-7} \cmidrule(lr){8-13} \cmidrule(lr){14-19}
Method & P* & Otsu & SEL26 & CC & CF & BGB & P* & Otsu & SEL26 & CC & CF & BGB & P* & Otsu & SEL26 & CC & CF & BGB \\
\midrule
SCLIP & 35.91 & 49.79 & 57.61 & 52.63 & 53.45 & 60.76 & 21.68 & 30.26 & 33.16 & 31.78 & 32.41 & 33.07 & 30.41 & -- & 23.47 & -- & 25.19 & 31.93 \\
ClearCLIP & 37.77 & 47.65 & 58.43 & 56.40 & 56.90 & 51.80 & 23.32 & 30.59 & 34.36 & 35.01 & 35.58 & 33.04 & 31.45 & -- & 24.45 & -- & 28.76 & 32.57 \\
NACLIP & 36.79 & 45.48 & 55.89 & 50.37 & 51.80 & 57.02 & 21.55 & 28.52 & 32.97 & 33.07 & 33.48 & 32.34 & 30.65 & -- & 22.72 & -- & 27.44 & 31.89 \\
ResCLIP & 36.66 & 44.91 & 58.52 & 53.63 & 55.05 & 60.35 & 22.00 & 29.76 & 34.47 & 36.34 & 36.59 & 34.56 & 31.84 & -- & 24.18 & -- & 29.66 & 33.41 \\
SC-CLIP & 42.07 & 48.61 & 62.51 & 57.21 & 58.98 & 64.34 & 24.74 & 30.73 & 35.52 & 37.87 & 37.96 & 36.75 & 34.49 & -- & 25.69 & -- & 32.08 & 36.33 \\
ProxyCLIP & 40.57 & 50.33 & 64.09 & 59.51 & 60.61 & 61.27 & 23.89 & 32.40 & 37.07 & 39.10 & 39.35 & 36.69 & 34.48 & -- & 26.26 & -- & 32.09 & 35.74 \\
Trident & 40.02 & 50.77 & 66.05 & 62.24 & 63.22 & 67.08 & 24.90 & 33.44 & 38.38 & 41.42 & 41.58 & 41.29 & 36.17 & -- & 27.44 & -- & 33.59 & 38.03 \\
\bottomrule
\end{tabular}
\end{table*}

%% file: tables/tabA2_variants_full.tex
\begin{table}[pos=htb]
\caption{Derived variants with controlled background share: P*/BGB per method.}\label{tab:variants_full}
\centering\scriptsize\setlength{\tabcolsep}{3pt}
\begin{tabular}{@{}lccccc@{}}
\toprule
Method & Ctx 35\% & Ctx 53\% & Ctx 71\% & COCO 31\% & COCO 51\% \\
\midrule
SCLIP & 27.33/31.90 & 25.12/30.34 & 20.08/25.31 & 21.11/19.19 & 20.30/20.35 \\
ClearCLIP & 29.02/32.38 & 26.45/30.29 & 21.68/26.01 & 22.64/18.12 & 21.73/19.31 \\
NACLIP & 27.64/31.17 & 25.08/28.83 & 19.98/23.75 & 21.54/19.91 & 20.35/20.81 \\
ResCLIP & 28.72/32.54 & 26.14/29.88 & 21.14/24.58 & 22.67/22.53 & 21.34/23.31 \\
SC-CLIP & 31.19/36.20 & 28.41/34.14 & 23.00/29.13 & 23.97/22.56 & 22.70/23.66 \\
ProxyCLIP & 31.24/36.18 & 28.19/34.35 & 22.71/29.88 & 24.99/19.67 & 23.46/21.15 \\
Trident & 32.48/38.23 & 29.20/36.00 & 23.72/31.10 & 26.06/22.89 & 24.44/24.48 \\
\bottomrule
\end{tabular}
\end{table}

%% file: tables/tab5_decomposition_original.tex
\begin{table*}[pos=htb]
\caption{Where the official background gain comes from on the original datasets. THR: P* plus only the numeric part of the official trick ($\tau$, $s$, $a$); BGB: THR plus the official class-name file. $\Delta_{\mathrm{thr}}=$THR$-$P*, $\Delta_{\mathrm{name}}=$BGB$-$THR; $\Delta_{\mathrm{voc}}$: gain from removing background words that collide with a query class name (``bed'' on COCO-Object; VOC21 has none). Threshold share $=\overline{\Delta_{\mathrm{thr}}}/\overline{\text{BGB}-\text{P*}}$.}\label{tab:decomp}
\centering\scriptsize\setlength{\tabcolsep}{3pt}
\begin{tabular}{@{}lccccccccccc@{}}
\toprule
 & \multicolumn{5}{c}{VOC21} & \multicolumn{6}{c}{COCO-Object} \\
\cmidrule(lr){2-6} \cmidrule(lr){7-12}
Method & P* & THR & BGB & $\Delta_{\mathrm{thr}}$ & $\Delta_{\mathrm{name}}$ & P* & THR & BGB & $\Delta_{\mathrm{thr}}$ & $\Delta_{\mathrm{name}}$ & $\Delta_{\mathrm{voc}}$ \\
\midrule
SCLIP & 35.91 & 38.25 & 60.76 & $+$2.34 & $+$22.51 & 21.68 & 29.72 & 33.07 & $+$8.04 & $+$3.35 & $+$0.61 \\
ClearCLIP & 37.77 & 51.80 & 51.80 & $+$14.03 & $+$0.00 & 23.32 & 33.04 & 33.04 & $+$9.72 & $+$0.00 & -- \\
NACLIP & 36.79 & 37.00 & 57.02 & $+$0.21 & $+$20.02 & 21.55 & 27.72 & 32.34 & $+$6.17 & $+$4.62 & $+$0.57 \\
ResCLIP & 36.66 & 36.84 & 60.35 & $+$0.18 & $+$23.51 & 22.00 & 28.10 & 34.56 & $+$6.10 & $+$6.46 & $+$0.58 \\
SC-CLIP & 42.07 & 43.99 & 64.34 & $+$1.92 & $+$20.35 & 24.74 & 33.87 & 36.75 & $+$9.13 & $+$2.88 & $+$0.61 \\
ProxyCLIP & 40.57 & 48.80 & 61.27 & $+$8.23 & $+$12.47 & 23.89 & 36.21 & 36.69 & $+$12.32 & $+$0.48 & $+$0.71 \\
Trident & 40.02 & 47.96 & 67.08 & $+$7.94 & $+$19.12 & 24.90 & 39.08 & 41.29 & $+$14.18 & $+$2.21 & $+$0.73 \\
\midrule Mean & & & & $+$4.98 & $+$16.85 & & & & $+$9.38 & $+$2.86 & $+$0.54 \\
Threshold share & \multicolumn{5}{c}{0.23} & \multicolumn{6}{c}{0.77} \\
\bottomrule
\end{tabular}
\end{table*}